\documentclass[twoside,11pt]{article}
\usepackage{subcaption}
\usepackage[nohyperref]{jmlr2e}
\usepackage[utf8]{inputenc} % allow utf-8 input
\usepackage[T1]{fontenc}    % use 8-bit T1 fonts
\usepackage[dvipsnames]{xcolor}
\usepackage{url}
\usepackage{booktabs}
\usepackage{array}
\usepackage{multirow}
\usepackage{amsfonts,amsmath}
\usepackage{siunitx}
\usepackage{etoolbox}
\usepackage[inline]{enumitem}
\usepackage{microtype}
\usepackage{graphicx}
\usepackage{tikz}
\usepackage[colorlinks,linkcolor=blue,citecolor=blue,urlcolor=blue]{hyperref}
\usepackage[capitalise]{cleveref}

\newcommand{\sii}{S_{11}}
\newcommand{\sij}{S_{12}}
\newcommand{\sjj}{S_{22}}
\newcommand{\mi}{\bar{\imath}}
\newcommand{\mj}{\bar{\jmath}}

\title{Morpho-MNIST: Quantitative Assessment and Diagnostics \\ for Representation Learning}
\author{%
\name Daniel C.~Castro \email dc315@imperial.ac.uk \\[\interauthorskip]
\name Jeremy Tan \email jht16@imperial.ac.uk \\[\interauthorskip]
\name Bernhard Kainz \email b.kainz@imperial.ac.uk \\
\addr Biomedical Image Analysis Group \\
Imperial College London \\
London SW7 2AZ, United Kingdom
\AND
\name Ender Konukoglu \email ender.konukoglu@vision.ee.ethz.ch \\
\addr Computer Vision Laboratory \\
ETH Z\"urich \\
8092 Z\"urich, Switzerland
\AND
\name Ben Glocker \email b.glocker@imperial.ac.uk \\
\addr Biomedical Image Analysis Group \\
Imperial College London \\
London SW7 2AZ, United Kingdom
}

\editor{Isabelle Guyon}

\usepackage{lastpage}
\jmlrheading{20}{2019}{1-\pageref{LastPage}}{1/19; Revised 6/19}{10/19}{19-033}{Daniel Coelho de Castro, Jeremy Tan, Bernhard Kainz, Ender Konukoglu, and Ben Glocker}
\ShortHeadings{Morpho-MNIST}{Castro, Tan, Kainz, Konukoglu, and Glocker}

\begin{document}

\maketitle

\begin{abstract}
Revealing latent structure in data is an active field of research, having introduced exciting technologies such as variational autoencoders and adversarial networks, and is essential to push machine learning towards unsupervised knowledge discovery. However, a major challenge is the lack of suitable benchmarks for an objective and quantitative evaluation of learned representations. 
To address this issue we introduce Morpho-MNIST, a framework that aims to answer: ``to what extent has my model learned to represent specific factors of variation in the data?” We extend the popular MNIST dataset by adding a morphometric analysis enabling quantitative comparison of trained models, identification of the roles of latent variables, and characterisation of sample diversity. We further propose a set of quantifiable perturbations to assess the performance of unsupervised and supervised methods on challenging tasks such as outlier detection and domain adaptation.
Data and code are available at \url{https://github.com/dccastro/Morpho-MNIST}.
\end{abstract}


\section{Introduction}\label{sec:introduction}

A key factor for progress in machine learning has been the availability of well curated, easy-to-use, standardised and sufficiently large annotated datasets for benchmarking different algorithms and models. This has led to major advances in speech recognition, computer vision, and natural language processing. A commonality between these tasks is their natural formulation as supervised learning tasks, wherein performance can be measured in terms of accuracy on a test set.

The general problem of representation learning (i.e.\ to reveal latent structure in data) is more difficult to assess due the lack of suitable benchmarks. Although the field is very active, with many recently proposed techniques such as probabilistic autoencoders and adversarial learning, it is less clear where the field stands in terms of progress or which approaches are more expressive for specific tasks. The lack of reproducible ways to quantify performance has led to subjective means of evaluation: visualisation techniques have been used to show low-dimensional projections of the latent space and visual inspection of generated or reconstructed samples are popular to provide subjective measures of descriptiveness. On the other hand, the quality of sampled images generally tells us little about how well the learned representations capture known factors of variation in the training distribution. In order to advance progress, the availability of tools for objective assessment of representation learning methods seems essential yet lacking.

This paper introduces Morpho-MNIST, a collection of shape metrics and perturbations, in a step towards quantitative assessment of representation learning. We build upon one of the most popular machine learning benchmarks, MNIST, which despite its shortcomings remains widely used. While MNIST was originally constructed to facilitate research in image classification, in the form of recognising handwritten digits \citep{LeCun1998}, it has found its use in representation learning, for example, to demonstrate that the learned latent space yields clusters consistent with digit labels. Methods aiming to disentangle the latent space claim success if individual latent variables capture specific style variations (e.g.\ stroke thickness, sidewards leaning digits and other visual characteristics).

It is important to emphasise that MNIST is a prime representative for datasets of rasterised two-dimensional shapes. In this type of data, while the underlying shapes may have relatively simple and well understood factors of variation, they manifest in the raster images as high-order correlations between large numbers of pixels. The challenge for a learning agent is to abstract away the low-level pixel intensity variations and recover a meaningful shape representation. MNIST is one of the interesting cases wherein it is possible to explicitly measure morphological attributes from the images themselves, enabling us to directly evaluate to what extent a trained model has learned to represent them.

% The main appeal of selecting MNIST as a benchmark for representation learning is that, while manifesting complex interactions between pixel intensities and underlying shapes, it has well understood and easily measurable factors of variation.
More generally, MNIST remains popular in practice due to several factors: it allows reproducible comparisons with previous results reported in the literature; the dataset is sufficiently large for its complexity and consists of small, two-dimensional greyscale images defining a tractable ten-class classification problem; computation and memory requirements are low; most popular deep learning frameworks and libraries offer tutorials using MNIST, which makes it straightforward for new researchers to enter the field and to experiment with new ideas and explore latest developments. Furthermore, there is now renewed interest in MNIST due to the recent replication of the original generation pipeline and the rediscovery of additional 50,000 test images, beckoning researchers to re-evaluate over 20 years of research on MNIST \citep{Yadav2019}. We leverage these qualities and extend MNIST in multiple ways, as summarised in the following.%\todo{Mention renewed interest in MNIST \citep{Yadav2019}}

% Besides making use of MNIST to evaluate supervised classification methods, researchers have found creative ways of using it for analysing unsupervised methods, as outlined above. To benefit from its qualities and to promote wide adoption of our proposed quantitative assessment of representation learning, we extend MNIST in multiple ways, as summarised in the following.

\subsection{Contributions}

% Our aim is to bridge the gap between methodology-focused research and critical real-world applications that could benefit from latest machine learning methods. 
Our aim is to provide a useful resource to the machine learning community, opening a variety of new avenues for experimentation and analysis.
As we preserve the general properties of MNIST---such as image size, file format, numbers of training and test images, and the original ten-class classification problem---we hope this new quantitative framework for assessing representation learning will experience widespread adoption and may inspire further extensions facilitated by a publicly available Morpho-MNIST code base.

\begin{figure}
	\centering
	\raisebox{-.5\height}{\includegraphics[trim={0 2cm 0 0},clip,scale=.27]{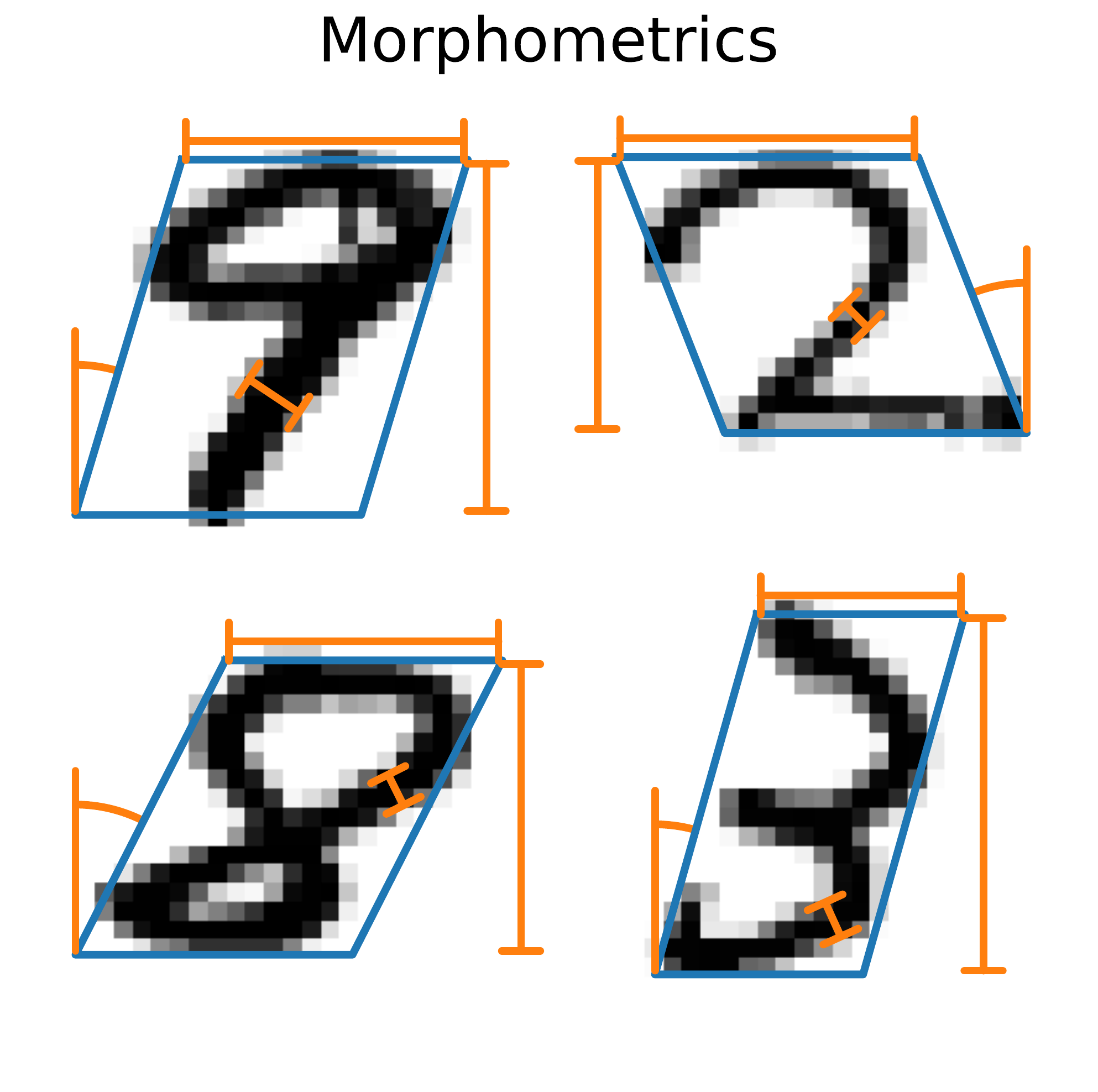}}
    \hfill
    \raisebox{-.5\height}{\includegraphics[scale=.54]{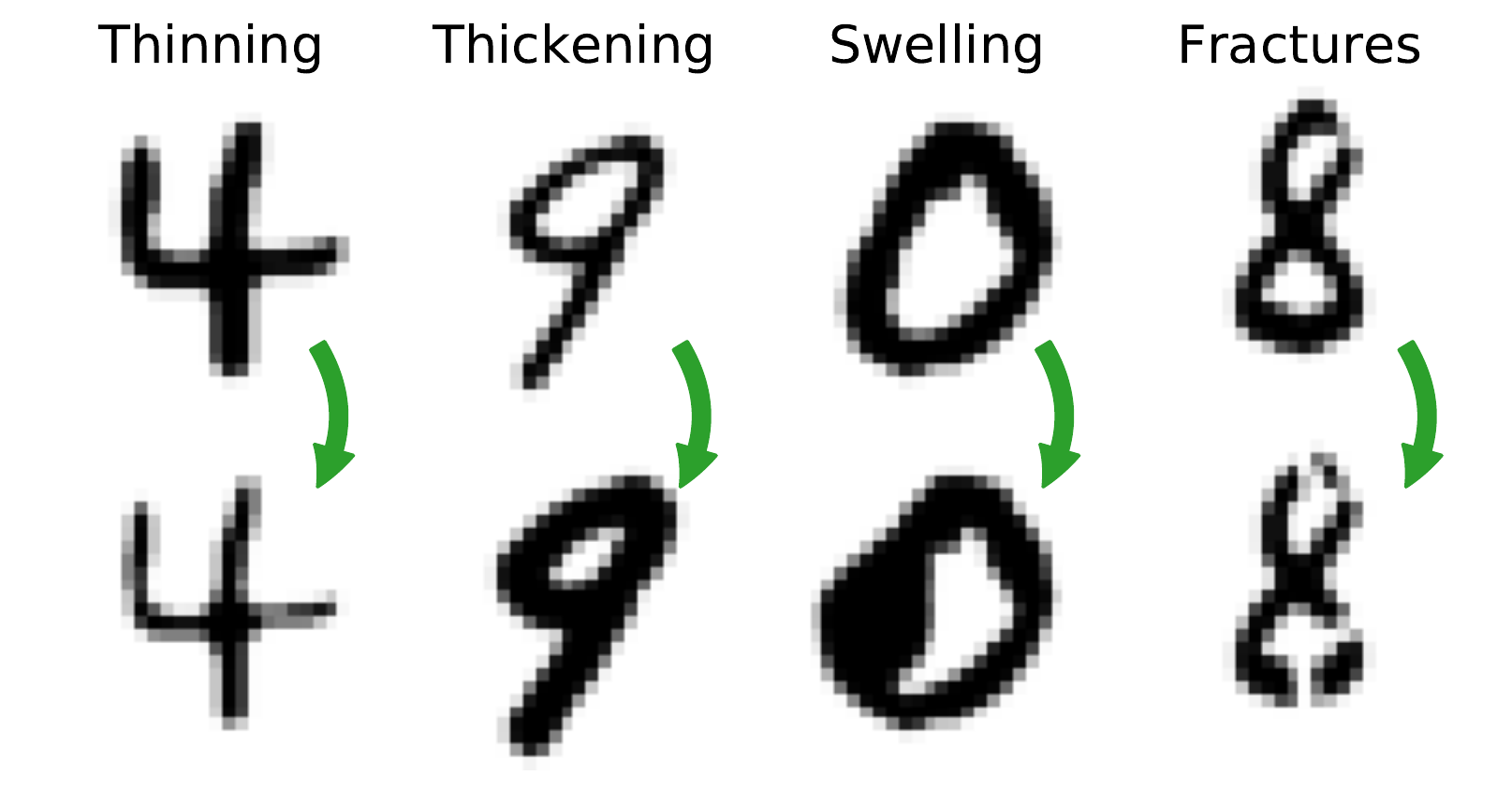}}
	\caption{\emph{Left:} MNIST morphometrics---stroke thickness and length (not shown), width, height and slant of digits. \emph{Right:} MNIST perturbations (many more examples of each type in \cref{app:pert_examples}).}
    \label{fig:morpho-mnist}
\end{figure}

% In this work, we augment MNIST with shape attributes, like the ones in the annotated datasets mentioned further below, enabling quantitative evaluation of inferred representations with this ubiquitous dataset. Moreover, these tools can be used to measure model \emph{samples}, enabling assessment of \emph{generative} performance, in terms of sample diversity and disentanglement of latent variables.

\subsubsection{Morphometrics}
We propose to describe real and generated digit images in terms of measurable shape attributes. These include stroke thickness and length, and the width, height, and slant of digits (see \cref{fig:morpho-mnist}, left). Whereas some of these MNIST properties have been analysed only visually in previous work, we demonstrate that quantifying each of them allows us to objectively characterise the roles of inferred representations.
% Moreover, these tools can be used to measure model samples, enabling assessment of \emph{generative} performance with respect to sample diversity (\cref{sec:diversity}) and disentanglement of latent variables (\cref{sec:disentanglement}).

% A crucial limitation of approaches for evaluating learned representations has been their dependence on \emph{annotated synthetic datasets} \citep{Eastwood2018}, such as the ones listed in \cref{sec:related_annotated}. This restricts their application to the analysis of \emph{inferential} behaviour (data $\to$ representation), even when dealing with generative models. In \cref{sec:disentanglement_generative} we illustrate how the measurement of data attributes enables investigation of \emph{generative} performance (representation $\to$ data) with respect to sample diversity (\cref{sec:diversity}) and disentanglement of latent variables (\cref{sec:disentanglement}).

More generally, the exclusive reliance on synthetic annotated datasets (\cref{sec:related_annotated}) has been a crucial limitation for research into evaluation of learned representations \citep{Eastwood2018}, as this restricts us to the analysis of \emph{inferential} behaviour (data $\to$ representation), even when studying generative models. The tools we introduce here can be used to measure model samples, effectively enabling direct assessment of \emph{generative} performance (representation $\to$ data).
% with respect to sample diversity (\cref{sec:diversity}) and disentanglement of latent variables (\cref{sec:disentanglement}).

These measurements can be directly employed to re-evaluate existing models and may be added retrospectively to previous experiments involving the original MNIST dataset. Adoption of our morphometric analysis may provide new insights into the effectiveness of representation learning methods in terms of revealing meaningful latent structures. Furthermore, for other datasets it suffices to design the relevant scalar metrics and include them in the very same evaluation framework.

\subsubsection{Perturbations}
We introduce a set of parametrisable global and local perturbations, inspired by natural and pathological variability in medical images. Global changes involve overall thinning and thickening of digits, while local changes include both swelling and fractures (see examples on the right in \cref{fig:morpho-mnist} and many more in \cref{app:pert_examples}). The perturbation framework we present in \cref{sec:perturbations}---illustrating various ways to exploit the outputs of the proposed processing pipeline (e.g.\ skeleton, distance map, and even morphometrics themselves; see \cref{sec:pipeline})---is sufficiently general that additional transformations can easily be designed as necessary. Injecting these perturbations into the dataset adds a new type of complexity to the data manifold and opens up a variety of interesting applications.% enables assessment of unsupervised and supervised methods on challenging tasks such as outlier detection and domain adaptation.

The proposed perturbations are designed to enable a wide range of new studies and applications for both supervised and unsupervised tasks. Detection of `abnormalities' (i.e.\ local perturbations) is an evident application, although more challenging tasks can also be defined, such as classification from noisy/corrupted data, domain adaptation, localisation of perturbations, characterising semantics of learned latent representations, and more. We explore a few supplementary examples of supervised tasks in \cref{sec:supervised}.

% Our work is inspired by problems in the biomedical imaging domain. Traditional supervised image classification methods often cannot be applied directly because of extreme class imbalances in the available training data. A large amount of data is available from healthy volunteers but low prevalence of highly variable pathologies paired with clinically required detection rates beyond 95\%, ideally without false negatives, makes representation learning a more attractive choice for such domains. Accurate modelling of normal appearance is desired in medicine to narrow the focus of clinical personnel to pathological and anatomically challenging cases. Representation learning furthermore paves the way to fully automatic identification of unknown or unexpected abnormalities during investigation. In this context our approach loosely corresponds to modelling common pathological characteristics, such as atrophy, hypertrophy, tumours, and bone fractures.

\subsection{Related Work: Datasets}

In this section, we provide an overview of some datasets that are related to MNIST, by either sharing its original source content, containing transformations of the original MNIST images or being distributed in the same format for easy replacement. We also mention a few prevalent datasets of synthetic images with generative factor annotations, analogous to the morphometrics proposed in this paper.

\subsubsection{NIST Datasets}
The MNIST (modified NIST) dataset \citep{LeCun1998} was constructed from handwritten digits in NIST Special Databases 1 and 3, now released as Special Database 19 \citep{Grother2016}. \citet{Cohen2017} generated a much larger dataset based on the same NIST database, containing additional upper- and lower-case letters, called EMNIST (extended MNIST). \citeauthor{LeCun1998}'s original processing pipeline was recently replicated, enabling the rediscovery of 50,000 test images that had been excluded from MNIST---a dataset now released as QMNIST \citep{Yadav2019}.

\subsubsection{MNIST Perturbations}
The seminal paper by \citet{LeCun1998} employed data augmentation using planar affine transformations including translation, scaling, squeezing, and shearing. \citet{Loosli2007} employed random elastic deformations to construct the Infinite MNIST dataset. Other MNIST variations include rotations and insertion of random and structured background \citep{Larochelle2007}, and \citet{Tieleman2013} applied spatial affine transformations and provided ground-truth transformation parameters.

\subsubsection{MNIST Format}
Due to the ubiquity of the MNIST dataset in machine learning research and the resulting multitude of compatible model architectures available, it is appealing to release new datasets in the same format (28$\times$28, 8-bit grayscale images). One such effort is Fashion-MNIST \citep{Xiao2017}, containing images of clothing articles from ten distinct classes, adapted from an online shopping catalogue. Another example is notMNIST \citep{Bulatov2011}, a dataset of character glyphs for letters `A'--`J' (also ten classes), in a challengingly diverse collection of typefaces.

\subsubsection{Annotated Datasets}\label{sec:related_annotated}
Synthetic computer vision datasets that are popular for evaluating disentanglement of learned latent factors of variation include those from \citet{Paysan2009} and \citet{Aubry2014}. They contain 2D renderings of 3D faces and chairs, respectively, with ground-truth pose parameters (azimuth, elevation) and lighting conditions (faces only). A further initiative in that direction is the dSprites dataset \citep{Matthey2017}, which consists of binary images containing three types of shapes with varying location, orientation and size. The availability of the ground-truth values of such attributes has motivated the accelerated adoption of these datasets in the evaluation of representation learning algorithms.

\subsection{Related Work: Quantitative Evaluation}

Evaluation of representation learning is a notoriously challenging task and remains an open research problem. Numerous solutions have been proposed, with many of the earlier ones focusing on the test log-likelihood under the model \citep{Kingma2014} or, for likelihood-free models, under a kernel density estimate (KDE) of generated samples \citep{Goodfellow2014,Makhzani2016}---shown not to be reliable proxies for the true model likelihood \citep{Theis2016}.

Another perspective for evaluation of generative models of images is the visual fidelity of its samples to the training data, which would normally require manual inspection. To address this issue, a successful family of metrics have been proposed, based on visual features extracted by the Inception network \citep{Szegedy2016}. The original Inception score \citep{Salimans2016} relies on the `crispness' of class predictions, whereas the Fr\'echet Inception distance \citep[FID,][]{Heusel2017} and the kernel Inception distance \citep[KID,][]{Binkowski2018} statistically compare high-level representations instead of the final network outputs.

Although the approaches above can reveal vague signs of mode collapse, it may be useful to diagnose this phenomenon on its own. With this objective, \citet{Arora2018} proposed to estimate the support of the learned distribution (assumed discrete) using the birthday paradox test, by counting pairs of visual duplicates among model samples. Unfortunately, the adoption of this technique is hindered by its reliance on manual visual inspection to flag identical images. 

There have been several attempts at quantifying representation disentanglement performance. For example, \citet{Higgins2017} proposed to use the accuracy in predicting which factor of variation was held fixed in a simulated dataset. There exist further information-theoretic approaches, involving the KL divergence contribution from each latent dimension \citep{Dupont2018} or their mutual information with each known generative factor \citep{Chen2018}. Yet another method, explored in \citet{Kumar2018}, is based on the predictive accuracy of individual latent variables to each generative factor. The concurrent work of \citet{Eastwood2018} introduced a comprehensive methodology for characterising different aspects of representation disentanglement.

\section{Morphometry}

Meaningful morphometrics are instrumental in characterising distributions of rasterised shapes, such as MNIST digits, and can be useful as additional data for downstream learning tasks. We regard the current choice of morphometrics as a minimal useful collection of attributes that are largely independent and can be estimated robustly. Although further more intricate metrics can be defined, they are likely to be entangled with the ones defined here and to be noisy due to the limited resolution of the original images. We begin this section by describing the image processing pipeline employed for extracting the metrics and for applying perturbations (\cref{sec:perturbations}), followed by details on the computation of each measurement.

\subsection{Processing Pipeline}\label{sec:pipeline}

\begin{figure}[tb]
	\centering
    \includegraphics[width=\textwidth]{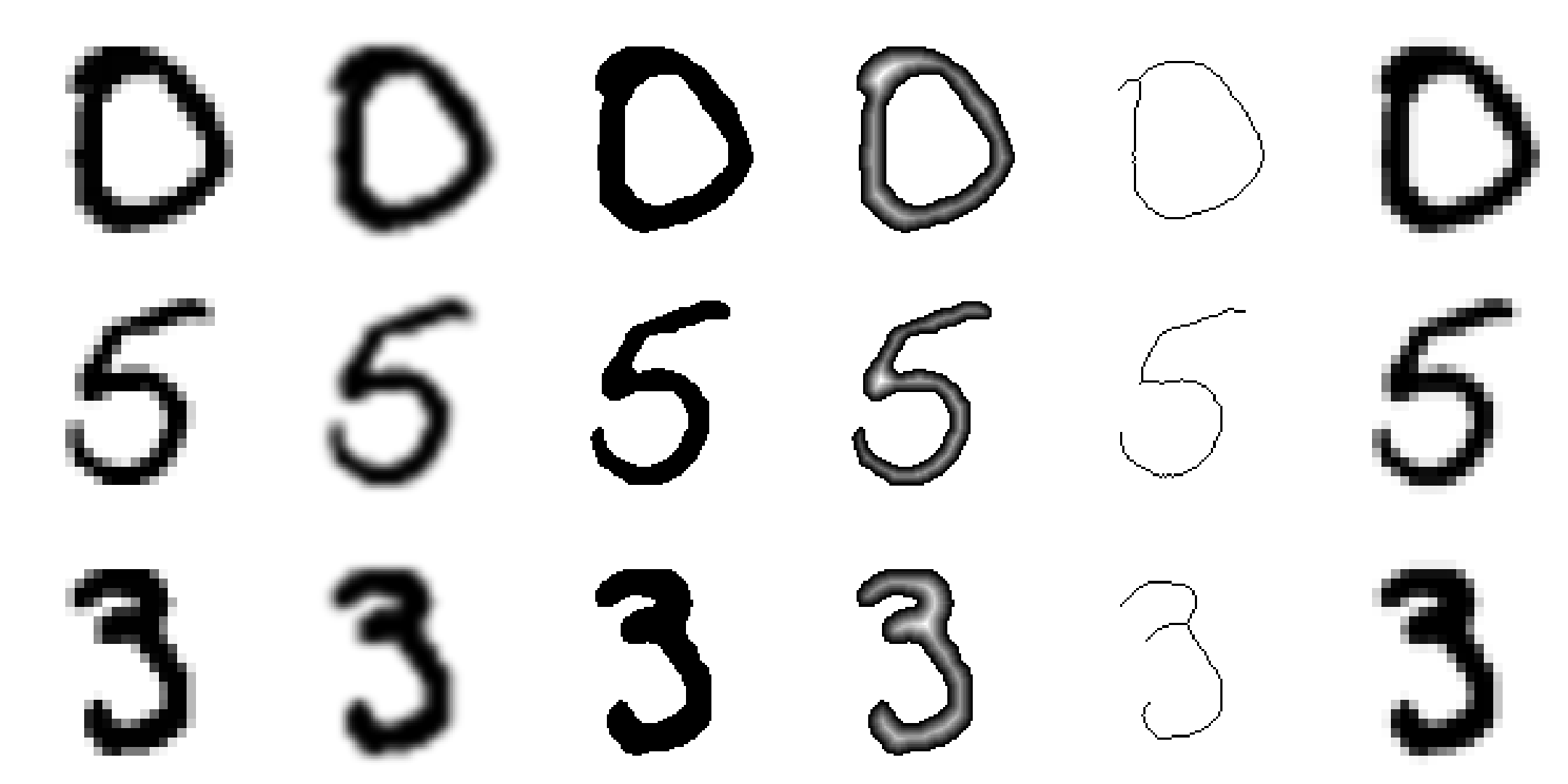}
    \caption{Stages of the image processing pipeline. \emph{Left to right:} original image, upscaled image, binarised image, distance transform, skeleton, downscaled image.}
    \label{fig:pipeline}
\end{figure}

The original 28$\times$28 resolution of the MNIST images is generally not high enough to enable satisfactory morphological processing: stroke properties (e.g.\ length, thickness) measured directly on the binarised images would likely be inaccurate and heavily quantised. To mitigate this issue and enable sub-pixel accuracy in the measurements, we propose to use the following processing steps:
\begin{enumerate}[itemsep=0pt]
    \item \emph{Upscale} (e.g.\ $\times$4, to 112$\times$112);\footnote{Up- and downscaling by a factor of $f$ are done with bicubic interpolation and Gaussian smoothing (bandwidth ${\sigma=2f/6}$), following \texttt{scikit-image} defaults \citep{VanderWalt2014}.}
    \item \emph{Binarise:} the blurry upscaled image is thresholded at half its intensity range (e.g.\ 128 for most images), which ensures thin or faint digits are not erased;
    \item Compute \emph{Euclidean distance transform} (EDT): each pixel within the digit boundaries contains the distance to its nearest boundary point;
    \item \emph{Skeletonise:} detect the ridges of the EDT, i.e.\ the locus of points equidistant to two or more boundary points (also known as the \emph{medial axis}) \citep{Blum1967};
    \item Optionally apply \emph{perturbation} to binarised image (cf.\ \cref{sec:perturbations});
    \item \emph{Downscale} binarised or perturbed image to original resolution.
\end{enumerate}%
% \begin{enumerate}[itemsep=0pt]
%     \item Upscale (e.g.\ $\times$4, to 112$\times$112)\footnote{Up- and downscaling by a factor of $f$ are done with bicubic interpolation and Gaussian smoothing (bandwidth ${\sigma=2f/6}$), following \texttt{scikit-image} defaults \citep{VanderWalt2014}.}
%     \item Binarise (e.g.\ threshold $\geq$128)
%     \item Compute Euclidean distance transform (EDT) from boundaries
%     \item Skeletonise (medial axis, i.e.\ ridges of EDT)
%     \item Apply perturbation (cf.\ \cref{sec:perturbations})
%     \item Downscale to original resolution
% \end{enumerate}

We illustrate the pipeline in \cref{fig:pipeline}. The binary high-resolution digits have smooth boundaries and faithfully capture subtle variations in contour shape and stroke thickness that are only vaguely discernible in the low-resolution images. Additionally, note how the final downscaled image is almost indistinguishable from the original.

\begin{figure}[tb]
	\centering
    \includegraphics[width=\textwidth]{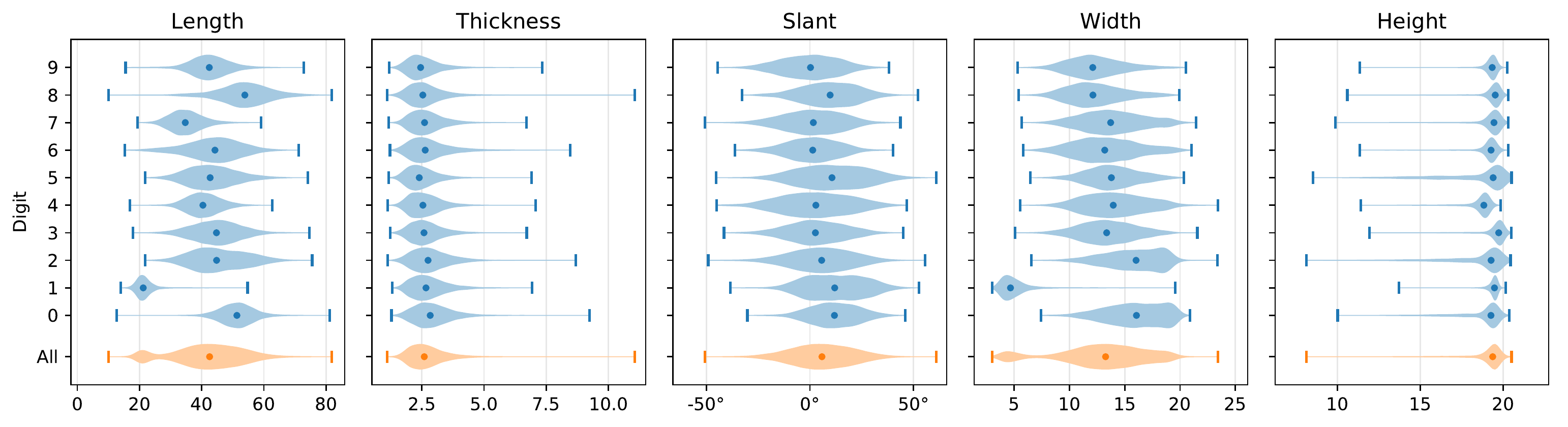}
    \caption{Distribution of morphological attributes in the plain MNIST training dataset}
    \label{fig:morpho_distribution}
\end{figure}

All morphometric attributes described below are calculated for each digit after applying steps 1--4 of this pipeline. The distributions for plain MNIST are plotted in \cref{fig:morpho_distribution}, and the distributions after applying each type of perturbation can be found in \cref{app:pert_morpho}.

\subsection{Stroke Length}

Here we approximate the trace of the pen tip, as a digit was being written, by the computed morphological skeleton. In this light, the total length of the skeleton is an estimate of the length of the pen stroke, which in turn is a measure of shape complexity.

It can be computed in a single pass by accumulating the Euclidean distance of each skeleton pixel to its immediate neighbours, taking care to only count the individual contributions once. This approach is more accurate and more robust against rotations than a na\"ive estimate by simply counting the pixels.

\subsection{Stroke Thickness}\label{sec:thickness}

A prominent factor of style variation in the MNIST digits is the overall thickness of the strokes, due to both legitimate differences in pen thickness and force applied, and also to the rescaling of the original NIST images by different factors.

We estimate it by exploiting both the computed distance transform and the skeleton. Since skeleton pixels are equidistant to the nearest boundaries and the value of the EDT at those locations corresponds to the local half-width \citep{Blum1967}, we take twice the mean of the EDT over all skeleton pixels as our global estimate for stroke thickness.

\begin{figure}
    \centering
    % \missingfigure{Illustrations of morphometrics: (a) skeleton trace for length; (b) zoom-in with medial disk for stroke thickness; (c) image moments for slant (centred vertical ellipse $\to$ slanted ellipse); (d) vertical and slanted calipers for width/height?}
    \begin{subfigure}{.24\textwidth}
        \centering
        \includegraphics[width=\textwidth]{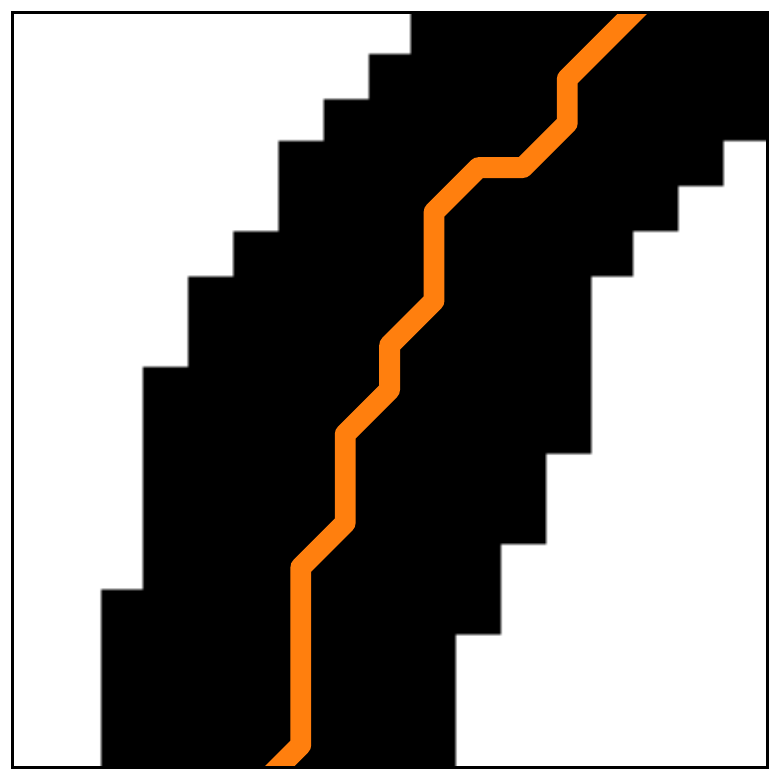}
    \end{subfigure}\hfill
    \begin{subfigure}{.24\textwidth}
        \centering
        \includegraphics[width=\textwidth]{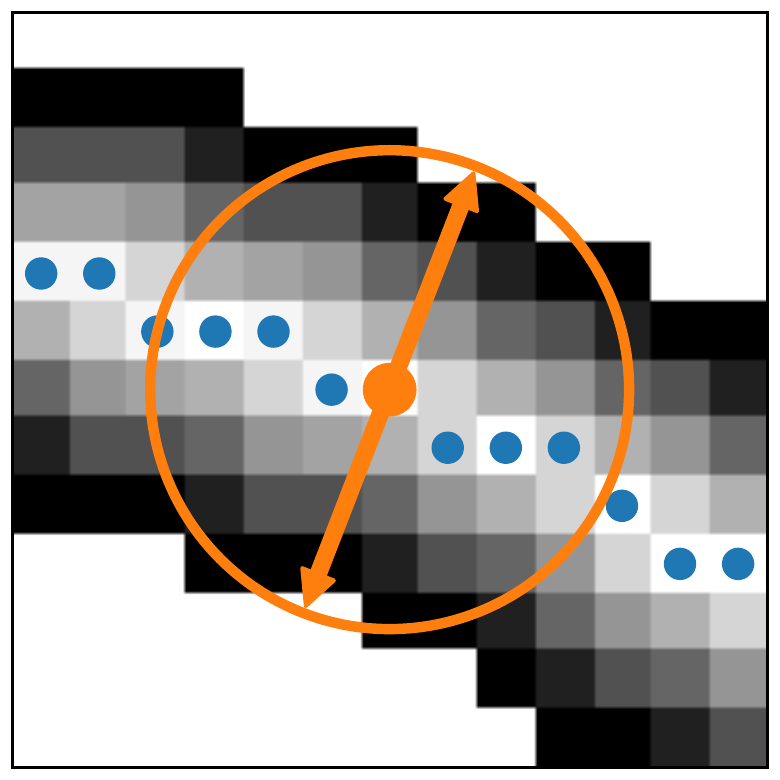}
    \end{subfigure}\hfill
    \begin{subfigure}{.24\textwidth}
        \centering
        \includegraphics[width=\textwidth]{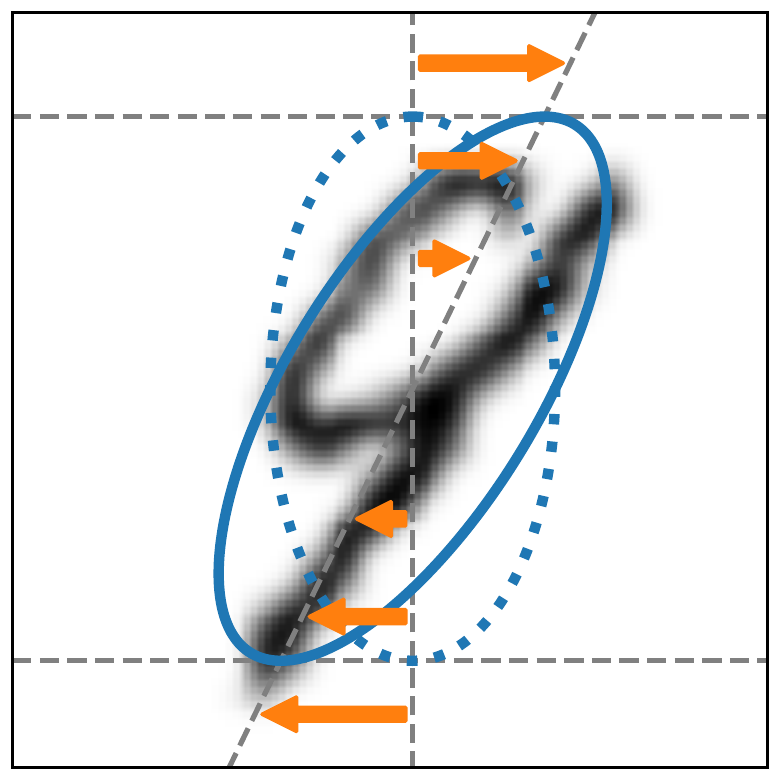}
    \end{subfigure}\hfill
    \begin{subfigure}{.24\textwidth}
        \centering
        \includegraphics[width=\textwidth]{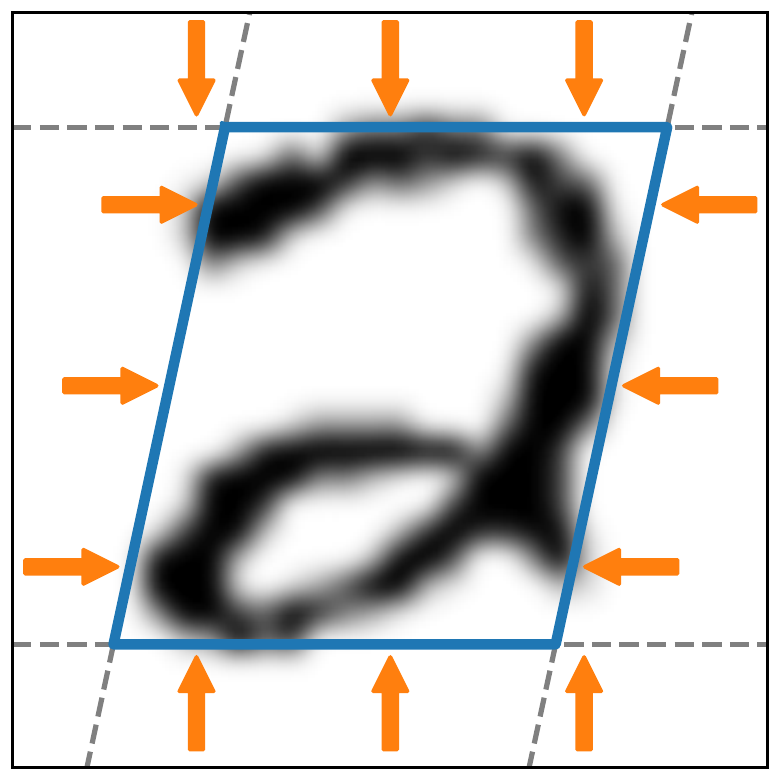}
    \end{subfigure}
    \caption{Measuring shape attributes. \emph{Left to right:} stroke length (length of binarised image's skeleton); stroke thickness (average distance transform over skeleton pixels $\times$2); slant (horizontal shearing angle); width and height (bounding parallelogram).}
\end{figure}

\subsection{Slant}\label{sec:slant}

The extent by which handwritten symbols lean right or left (forward and backward slant, respectively) is a further notorious and quantifiable dimension of handwriting style. It introduces so much variation in the appearance of characters in images that it is common practice in optical character recognition (OCR) systems to `deslant' them, in an attempt to reduce within-class variance \citep{LeCun1998,Teow2002}.

We adapt the referred deslanting methodology to describe the slant \emph{angle} of the handwritten digits. After estimating the second-order image moments,
\begin{equation}
    \mathbf{S} = \begin{pmatrix}
        \sum_{i,j} x_{ij} (i-\mi)^2 &
        \sum_{i,j} x_{ij} (i-\mi)(j-\mj) \\
        \sum_{i,j} x_{ij} (i-\mi)(j-\mj) &
        \sum_{i,j} x_{ij} (j-\mj)^2
    \end{pmatrix} ,
\end{equation}
we define the slant based on the horizontal shear:
\begin{equation}\label{eq:slant}
	\alpha = \arctan\!\left( -\frac{\sij}{\sjj} \right) ,
\end{equation}
where $x_{ij}$ is the intensity of the pixel at row $j$ and column $i$, and $(\mi,\mj)$ are the centroid coordinates. The minus sign ensures that positive and negative values of $\alpha$ correspond to forward and backward slant, respectively.

Note that $-\alpha$ is the angle by which the image would need to be horizontally sheared toward the vertical centre line such that the digit is upright (i.e.~principal axes aligned with the coordinate axes). Note how this differs from a \emph{rotation} angle, which would instead be defined in terms of $\frac12\arctan(2\sij/(\sjj-\sii))$.

\subsection{Width and Height}

It is useful to measure other general shape attributes, such as width, height, and aspect ratio, which also present substantial variation related to personal handwriting style.%
\footnote{Little variation in height is expected, since the original handwritten digits were scaled to fit a 20$\times$20 box \citep{LeCun1998}. Nevertheless, a minority of digits were originally wider than they were tall, which explains the long tails in the distribution of heights (\cref{fig:morpho_distribution}).}
To this end, we propose to fit a \emph{bounding parallelogram} to each digit, with horizontal and slanted sides (see \cref{fig:morpho-mnist}). We highlight that it does not suffice to simply fit a bounding box with perpendicular sides, since the width would clearly be confounded with slant. These metrics would therefore contain redundant information that obfuscates the true digit width.

The image is swept top-to-bottom with a horizontal boundary to compute a vertical marginal cumulative distribution function (CDF), and likewise left-to-right with a \emph{slanted} boundary for a horizontal marginal CDF, with angle $\alpha$ as computed above. The bounds are then chosen based on equal-tailed intervals containing a given proportion of the image mass---98\% in both directions (1\% from each side) proved accurate and robust in our experiments.

\section{Perturbations}\label{sec:perturbations}

% As discussed in \cref{sec:introduction}, we bring forward a number of parametrisable morphological perturbations to the MNIST digits, which we detail below.\todo{Rewrite}

As discussed in \cref{sec:introduction}, we bring forward a number of morphological perturbations for MNIST digits, to enable diverse applications and experimentation. In this section, we detail these parametrisable transformations, categorised as global or local.

\subsection{Global: Thinning and Thickening}

The first pair of transformations we present is based on simple morphological operations: the binarised image of a digit is dilated or eroded with a circular structuring element. Its radius is set proportionally to the estimated stroke thickness (\cref{sec:thickness}), so that the overall thickness of each digit will decrease or increase by an approximately fixed factor (here, -70\% and +100\%; see \cref{fig:examples_thin,fig:examples_thic}).

Since there is substantial thickness variability in the original MNIST data (cf.\ \cref{fig:morpho_distribution}) and most thinned and thickened digits look very plausible, we believe that these perturbations can constitute a powerful form of data augmentation for training. For the same reason, we have not included these perturbations in the abnormality detection experiments (\cref{sec:new_tasks}).%\todo{Revise this in line with updated supervised experiments}

\subsection{Local: Swelling}
{\renewcommand{\*}[1]{\mathbf{#1}}

In addition to the global transformations above, we introduce \emph{local} perturbations with variable location and extent, which are harder to detect automatically. Given a radius $R$, a centre location $\*r_0$ and a strength parameter $\gamma > 1$, the coordinates $\*r$ of pixels within distance $R$ of $\*r_0$ are nonlinearly warped according to a radial power transform:
\begin{equation}
	\*r \mapsto \*r_0 + (\*r-\*r_0) \Big( \frac{\lVert\*r-\*r_0\rVert}{R} \Big)^{\gamma-1} \,,
\end{equation}
leaving the remaining portions of the image untouched and resampling with bicubic interpolation.

In the experiments and released dataset, we set $\gamma=7$ and $R=3\sqrt{\theta}/2$, where $\theta$ is thickness. Unlike simple linear scaling with $\theta$, this choice for $R$ produces noticeable but not exaggerated effects across the thickness range observed in the dataset (cf.\ \cref{fig:examples_swel}). The centre location, $\*r_0$, is picked uniformly at random from the pixels along the estimated skeleton.
}

\subsection{Local: Fractures}

Finally, we describe the proposed procedure for adding fractures to an MNIST digit, where we define a fracture as a break in the continuity of a pen stroke. Because single fractures can in many cases be easily mistaken for true gaps between strokes, we add multiple fractures to each affected digit.

When selecting where to place a fracture, we attempt to avoid locations too close to stroke tips (points on the skeleton with a single neighbour) or fork points (more than two neighbours). This is achieved by sampling only among those skeleton pixels above a certain distance to these detected points. In addition, we desire fractures to be transversal to the pen strokes. Local orientation is determined based on second-order moments of the skeleton inside a window centred at the chosen location, and the length of the fracture is estimated from the boundary EDT. Finally, the fracture is drawn onto the high-resolution binary image with a circular brush along the estimated normal.%\footnote{As mentioned in \cref{sec:thickness}, the distances from the skeleton to either side are roughly equal. We add some tolerance by extending the line on both sides by 0.5\,px.}

In practice, we found that adding three fractures with 1.5\,px thickness, 2\,px minimum distance to tips and forks and angle window of 5$\times$5\,px$^2$ (`px' as measured in the low resolution image) produces detectable but not too obvious perturbations (see \cref{fig:examples_frac}). We also extend the lines on both ends by 0.5\,px to ensure some tolerance.

\section{Evaluation Case Studies}\label{sec:evaluation}

In this section, we demonstrate potential uses of the proposed framework: using morphometrics to characterise the distribution of samples from generative models and finding associations between learned latent representations and morphometric attributes. %In addition, we exemplify in \cref{sec:supervised} a variety of supervised tasks on the MNIST dataset augmented with perturbations.\todo{Update w.r.t.\ restructured \cref{sec:supervised}}
In addition, we exemplify in \cref{sec:supervised} a variety of novel ways to make use of morphometry and perturbations for evaluating predictive models on MNIST and beyond.

\subsection{Sample Diversity}\label{sec:diversity}

Here we aim to illustrate ways in which the proposed MNIST morphometrics may be used to visualise distributions learned by generative models and to quantify their agreement with the true data distribution in terms of these concrete attributes. Moreover, extracting such measurements from model samples could be a step toward diagnosing mode collapse.%\todo{Maybe move to future work, and/or discuss together with finding replicas?}

We exemplify this scenario with a generative adversarial network \citep[GAN,][]{Goodfellow2014} and a $\beta$-variational autoencoder \citep[$\beta$-VAE,][]{Higgins2017}, both with generator (resp.\ decoder) and discriminator architecture as used in the MNIST experiments in \citet{Chen2016}, and encoder mirroring the decoder. We train a $\beta$-VAE with $\beta=4$, as in \citet{Higgins2017}'s experiments on small binary images, and a vanilla GAN with non-saturating loss, both with 64-dimensional latent space. To explore the behaviour of a much less expressive model, we additionally train a GAN with only two latent dimensions.

\begin{figure}[p]
	\centering
	\begin{subfigure}[b]{.48\linewidth}
		\centering
		\includegraphics[width=\linewidth]{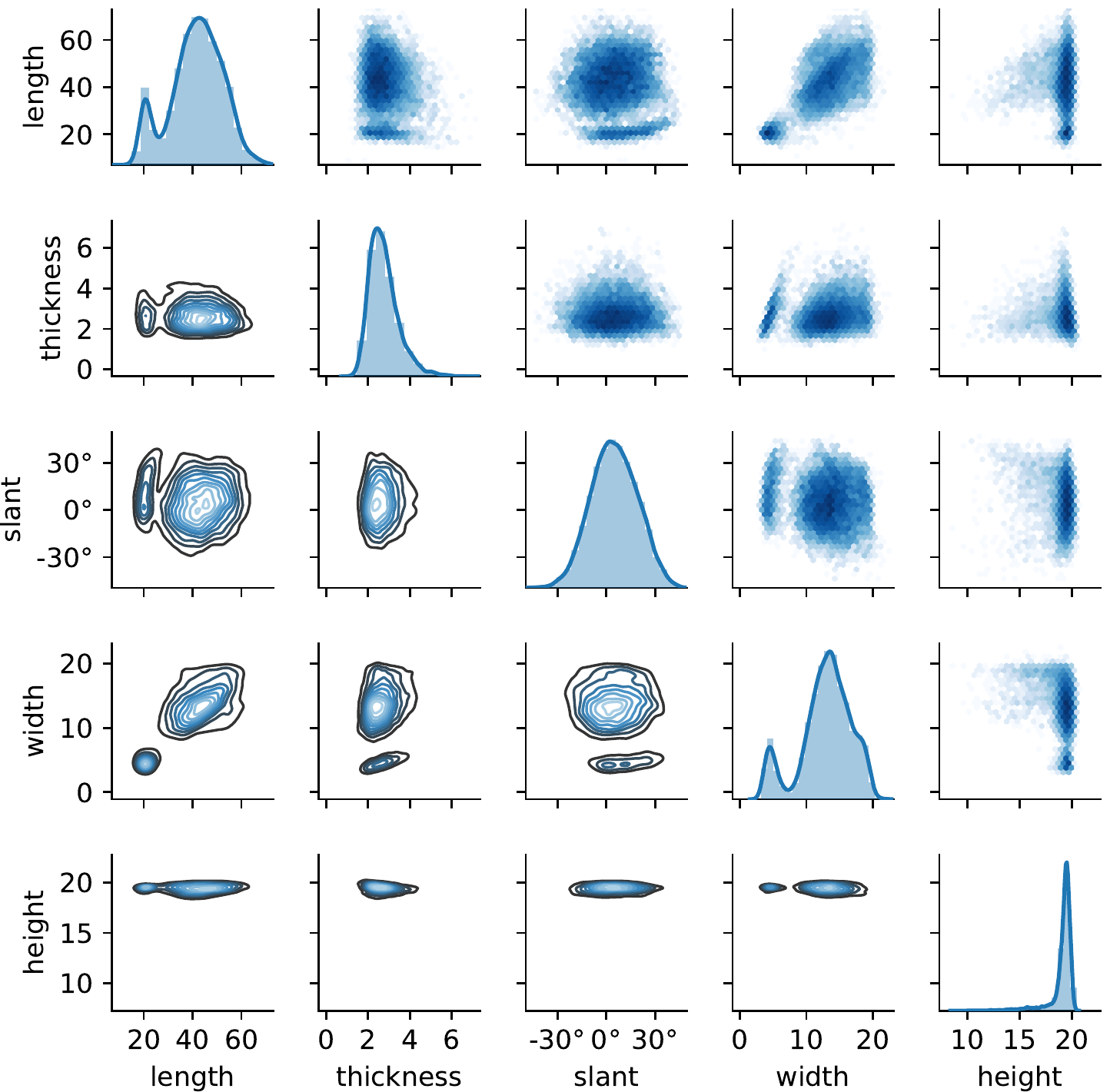}
		\caption{MNIST test images}
		\label{fig:sample_diversity_test}
	\end{subfigure} \hfill
	\begin{subfigure}[b]{.48\linewidth}
		\centering
		\includegraphics[width=\linewidth]{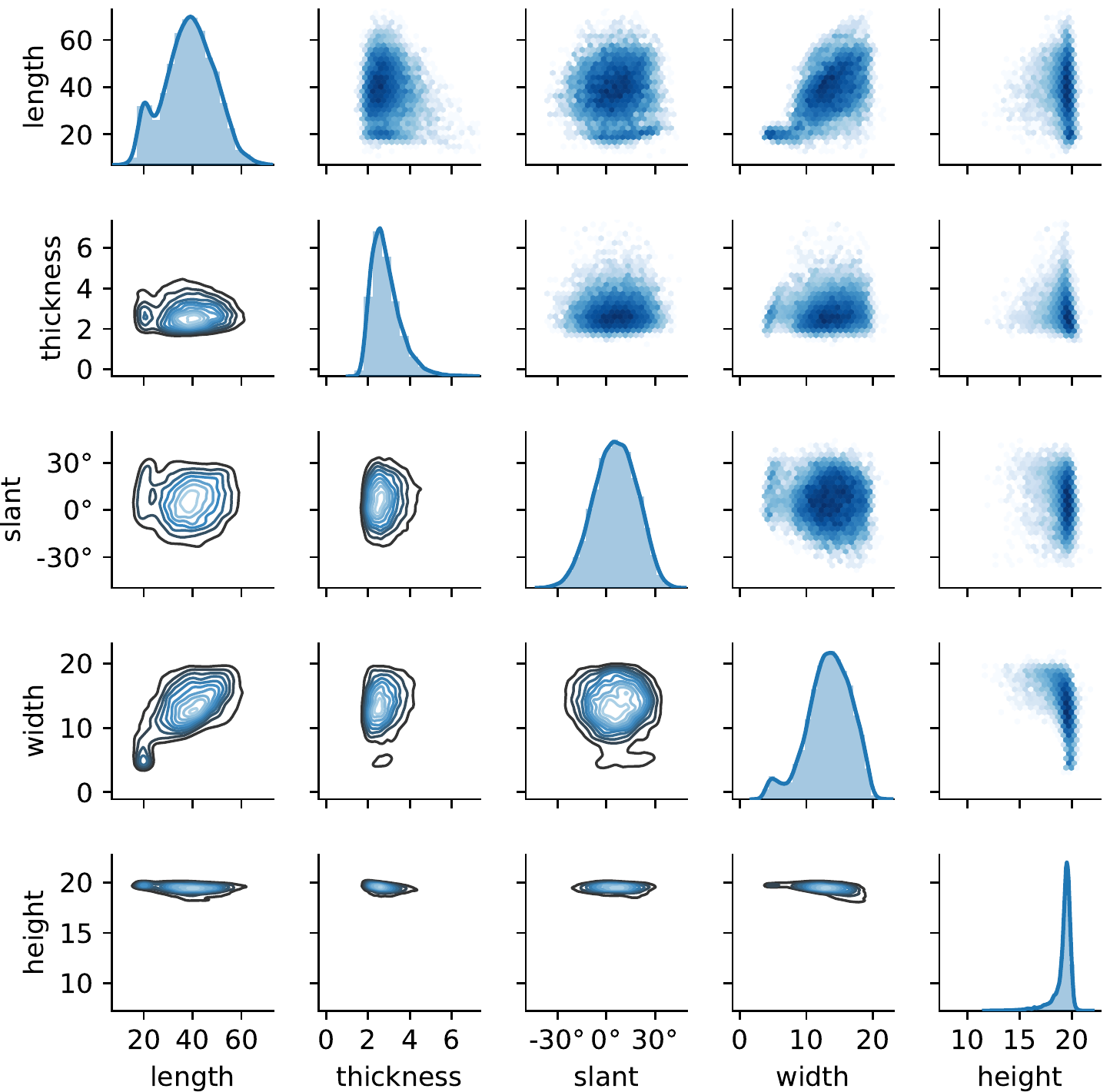}
		\caption{Samples from a $\beta$-VAE}
		\label{fig:sample_diversity_vae}
	\end{subfigure} \\[\baselineskip]
	\begin{subfigure}[b]{.48\linewidth}
		\centering
		\includegraphics[width=\linewidth]{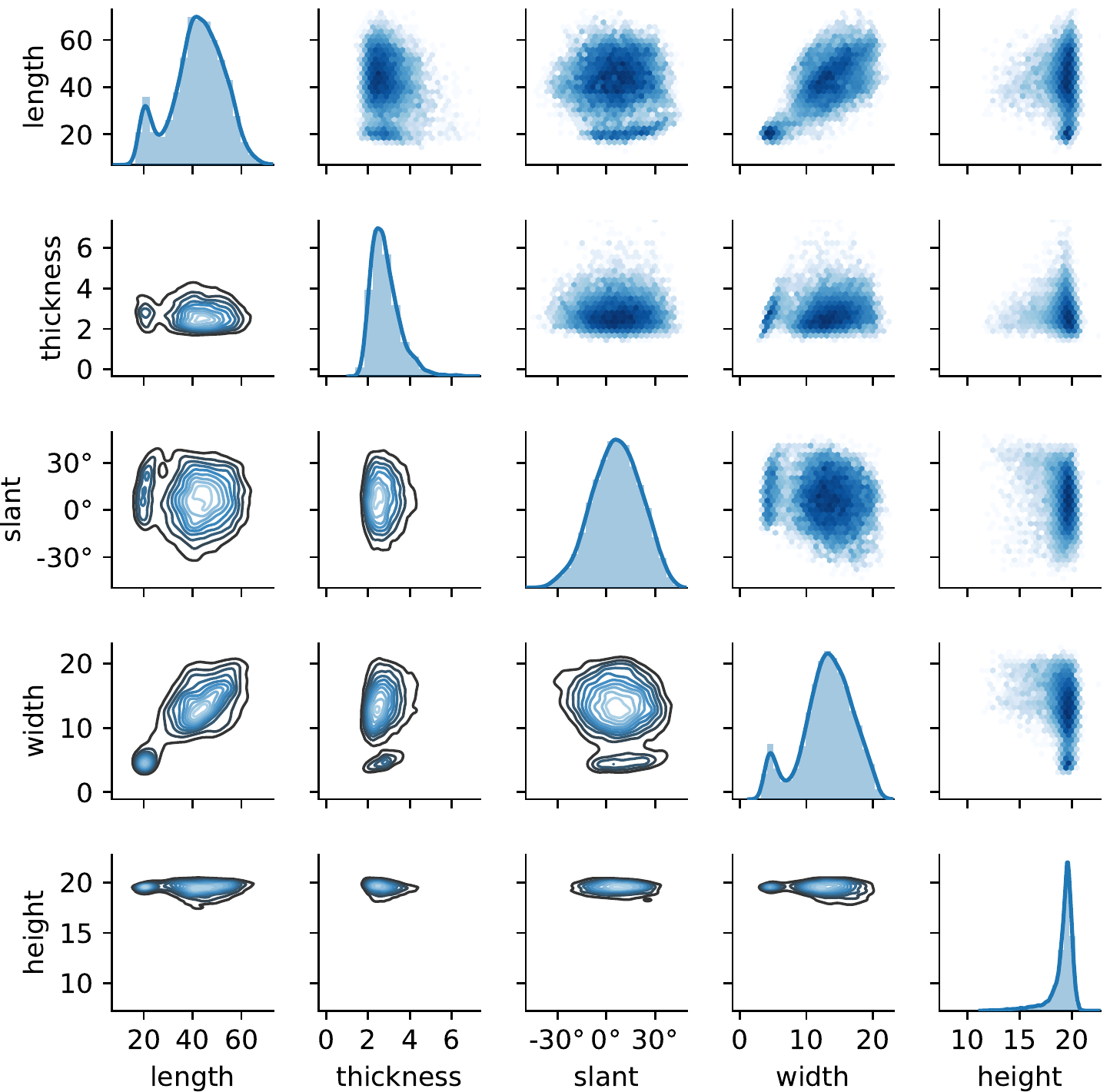}
		\caption{Samples from a GAN}
		\label{fig:sample_diversity_gan}
	\end{subfigure} \hfill
	\begin{subfigure}[b]{.48\linewidth}
		\centering
		\includegraphics[width=\linewidth]{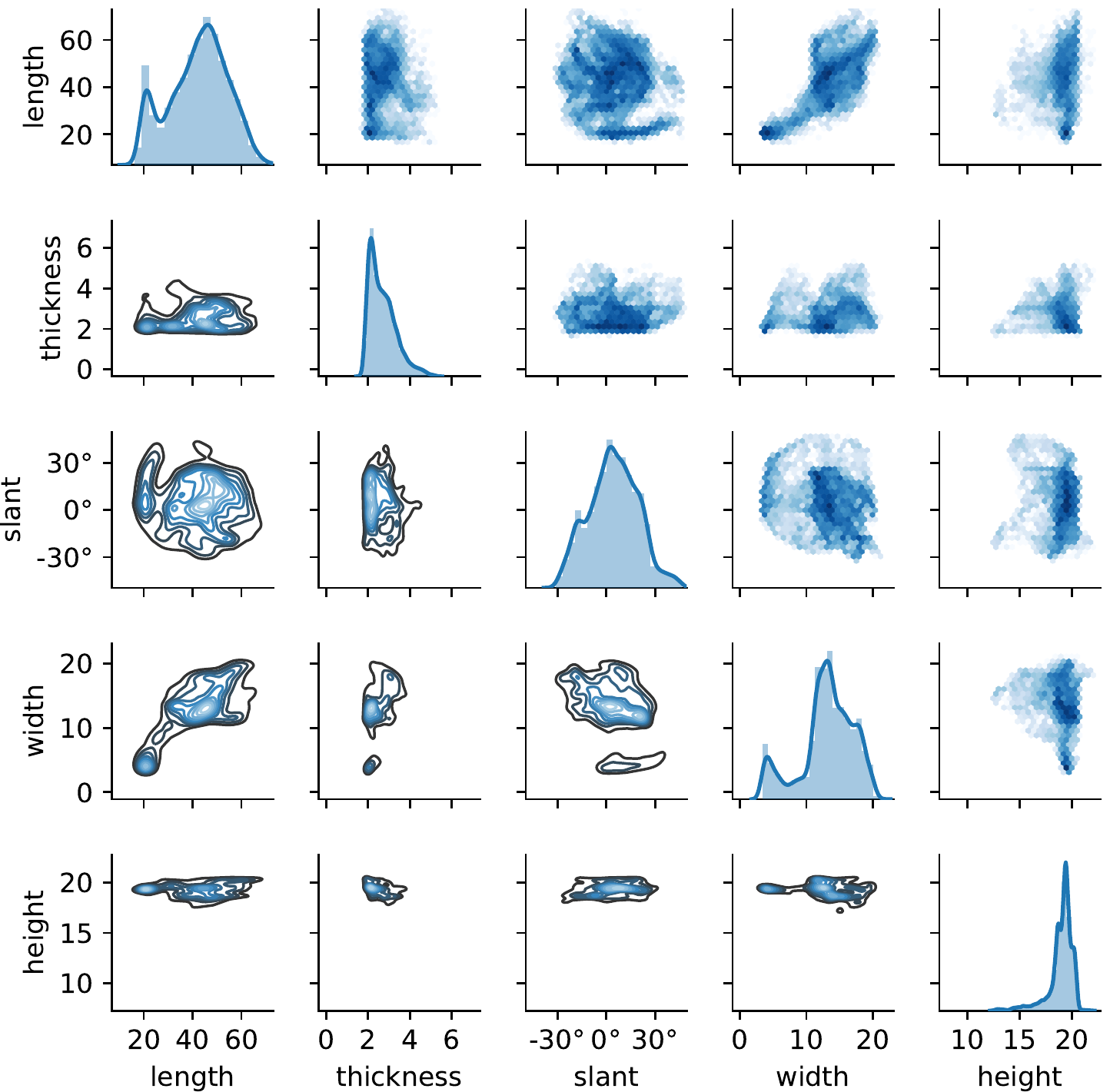}
		\caption{Samples from a low-dimensional GAN}
		\label{fig:sample_diversity_gan_collapse}
	\end{subfigure}
	\caption{Distribution of morphometric attributes for MNIST test dataset and samples from some generative models. Diagonals show marginal histograms and KDEs, upper-triangular plots show pairwise log-histograms and lower-triangular plots show pairwise KDEs.}
	\label{fig:sample_diversity}
\end{figure}

\subsubsection{Visualisation}
\Cref{fig:sample_diversity} illustrates the morphometric distributions of the plain MNIST test images and of 10,000 samples from each of these three models. As can be seen, morphometrics provide \emph{interpretable} low-dimensional statistics which allow comparing distributions learned by generative models between each other and with true datasets. While \cref{fig:sample_diversity_vae,fig:sample_diversity_gan} show model samples roughly as diverse as the true images, the samples from the low-dimensional GAN in \cref{fig:sample_diversity_gan_collapse} seem concentrated on certain regions, covering a distribution that is less faithful to the true one in \cref{fig:sample_diversity_test}.

\subsubsection{Statistical Comparison}
We argue that in this lower-dimensional space of morphometrics it is possible to statistically compare the distributions, since this was shown not to be effective directly in image space \citep[e.g.][]{Theis2016}. To this end, we propose to use kernel two-sample tests based on maximum mean discrepancy (MMD) between morphometrics of the test data and of each of the sample distributions \citep{Gretton2012}. The MMD statistic is a measure of dissimilarity between distributions, such that a value significantly greater than zero (small $p$-value) suggests a detectable difference. Here, we performed the linear-time asymptotic test described in \citet[\S 6]{Gretton2012} (details and further considerations in \cref{app:kernel}).

\begin{table}[tb]
	\centering
    \sisetup{table-number-alignment=center, mode=text, detect-weight, detect-inline-weight=math}
    \renewcommand{\bfseries}{\fontseries{b}\selectfont}
	\newrobustcmd{\B}{\bfseries}
    \begin{tabular}{%
    	lc
    	S[separate-uncertainty=true,
          table-format=1.3,
          table-figures-uncertainty=4]%
        S[add-integer-zero=false,
          table-format=0.4]}
    \toprule
    Test data vs. & Dims. & {$\operatorname{MMD}^2_l \pm \text{ std.\ error}$ ($\times 10^{-3}$)} & {$p$} \\
    \midrule
    $\beta$-VAE	& 64	& 0.792		\pm 1.569	& .3068 \\
    GAN			& 64	& 1.458		\pm 1.650	& .1885 \\
    GAN			& 2 	& \B 8.876	\pm 1.807	& \B .0000 \\
%     $\beta$-VAE	& 64	& 1.439		\pm 1.466	& .1632 \\
%     GAN			& 64	& 0.602		\pm 1.718	& .3631 \\
%     GAN			& 2 	& \B 7.400	\pm 1.798	& \B .0000 \\
    \bottomrule
    \end{tabular}
	\caption{Kernel two-sample tests between model samples and true test data. The bold row indicates a significant divergence, thus a failure of the model in faithfully reproducing the data distribution.}
    \label{tab:mmd_tests}
\end{table}

The results of these tests are displayed in \cref{tab:mmd_tests}, where we observe that samples from the high-dimensional $\beta$-VAE and GAN did not exhibit a significantly large MMD, thus there is no evidence of distributional mismatch. On the other hand, the low-dimensional GAN's samples show a significant departure from the data distribution, confirming the qualitative judgement based on comparing \cref{fig:sample_diversity_test,fig:sample_diversity_gan_collapse}.

% \subsubsection{Finding Replicas}
% One potentially fruitful suggestion would be to use a variant of hierarchical agglomerative clustering \added{\citep{Ward1963}} on sample morphometric attributes (e.g.\ using standardised Euclidean distance, or other suitable metrics). With a low enough distance threshold, it would be possible to identify groups of near-replicas, the abundance of which would signify mode collapse. Alternatively, this could be applicable as a heuristic in the birthday paradox test for estimating the support of the learned distribution \citep{Arora2018}.\todo{Merge this section into \cref{sec:discussion}}

\subsection{Disentanglement}\label{sec:disentanglement}

\newcommand{\InfoGANA}{\textsc{InfoGAN-A}}
\newcommand{\InfoGANB}{\textsc{InfoGAN-B}}
\newcommand{\InfoGANC}{\textsc{InfoGAN-C}}

\newcommand{\PlainData}{\textsc{plain}}
\newcommand{\GlobalData}{\textsc{global}}
\newcommand{\LocalData}{\textsc{local}}

% We argue that MNIST morphometrics not only allow the application of the InfoGAN's experimental set-up with dSprites to the MNIST dataset, but we can also evaluate the generative direction of such models, by correlating latent codes with morphometrics extracted from the corresponding sampled images.

In this experiment, we demonstrate that: (a) standard MNIST can be augmented with morphometric attributes to quantitatively study representations computed by an \emph{inference} model (as already possible with annotated datasets, e.g.\ dSprites and 3D faces); (b) we can measure shape attributes of samples to assess disentanglement of a \emph{generative} model, which is unprecedented to the best of our knowledge; and (c) this analysis can also diagnose when a model unexpectedly \emph{fails} to disentangle a known aspect of the data.

As for the sample diversity experiments in \cref{sec:diversity}, the purpose of this disentanglement analysis is to study the behaviour of specific trained model instances, and results may not generalise to the respective model classes or architectures, to different datasets, or even to distinct runs of the same set-up. Similarly to any other evaluation methodology, findings concerning disentanglement patterns are reproducible only to the extent that the model under study can be reliably retrained to a similar state. %\red{This is evidently the case with any evaluation methodology, although this caveat is often overlooked in the literature.}

\subsubsection{Methodology}\label{sec:disentanglement_methodology}
Using an approach related to the disentanglement measure introduced by \citet{Kumar2018}, we study the correlation structures between known generative factors and learned latent codes. Specifically, we compute the \emph{partial correlation} ($r_{c_i y_j \,\cdot\, \mathbf{c}_{-i}}$) between each latent code variable ($c_i$) and each morphometric attribute ($y_j$), controlling for the variation in the remaining latent variables ($\mathbf{c}_{-i}$).%
\footnote{For the categorical code, $c_1$, we take a single binary dummy variable for each category, $c_1^{(k)}$, while controlling only for the remaining codes ($c_2$, $c_3$ etc.) to avoid multicollinearity.}
As opposed to the simple correlation, this technique allows us to study the \emph{net} first-order effect of each latent code, all else being equal.

This partial correlation is calculated as follows, ranging over codes $i$ and attributes $j$:
\begin{equation}\label{eq:partial_corr}
    % \mathbf{P}^{(j)} = \begin{pmatrix}
    %     s^2_{y_j}   & s_{y_j c_1}   & \cdots    & s_{y_j c_D} \\
    %     s_{y_j c_1} & s^2_{c_1}     & \cdots    & s_{c_1 c_D} \\
    %     \vdots      & \vdots        & \ddots    & \vdots \\
    %     s_{y_j c_D} & s_{c_1 c_D}   & \cdots    & s^2_{c_D} \\
    % \end{pmatrix}^{-1} ,\quad
    r_{c_i y_j \,\cdot\, \mathbf{c}_{-i}} = -\frac{P^{(j)}_{1,i+1}}{\sqrt{P^{(j)}_{11} P^{(j)}_{i+1,i+1}}} \quad\in [-1,1] \,,
\end{equation}
where $\mathbf{P}^{(j)}$ is the sample precision matrix (i.e.\ inverse of sample covariance) of $(y_j, c_1, \dots, c_D)$ over all data points. Note that \cref{eq:partial_corr} is closely related to the least-squares coefficients for multiple linear regression of $y_j$ on $(c_1,\dots,c_D)$, and its magnitude can be directly interpreted as a relative importance score within the framework of \citet{Eastwood2018}.

We plot the calculated partial correlations in the form of tables, where each row corresponds to a shape attribute, and columns represent the latent variables. These tables should be read row-wise, as the scores are normalised separately for each shape factor. Entries close to zero do not indicate that the latent dimension is unrelated to the measured attribute, but rather that it is entangled with others and has little marginal effect on its own.

In addition, we demonstrate how our morphometry framework is fully compatible with contemporary disentanglement metrics by also computing the mutual information gap \citep[MIG score;][]{Chen2018} in each scenario. %For each generative factor (here, shape attribute), this score takes the mutual information with each latent code, then reports the difference between the top two.
This score is based on the mutual information of each pair of generative factor (here, shape attribute) and latent code dimension, and reports the gap between the two leading codes for each factor, normalised to $[0,1]$.
% \footnote{We compute mutual information for continuous variables by first discretising them into bins.}
It therefore quantifies how prominently the attribute is represented by a single latent dimension, i.e.\ its \emph{completeness} in the terminology of \citet{Eastwood2018}. The overall MIG for each model is given by the average over all attributes.

For these experiments, we trained InfoGAN \citep{Chen2016} instances with various latent space configurations, then took MAP estimates of latent codes for each image (i.e.\ maximal logit for categorical codes and mean for continuous codes) as predicted by the variational recognition network. Models were trained for 20 epochs using 64 images per batch, with no hyperparameter tuning. We emphasise that our goal was to illustrate how the proposed morphometrics can serve as tools to better understand whether models behave as intended, and not to optimally train them in each scenario.

% \todo[inline,caption={}]{Compute MIG \citep{Chen2018}?

% \begin{tabular}{lcc}
%     \toprule
%                 & Test      & Samples \\
%     \midrule
%     \InfoGANA   & 0.0582    & 0.0641 \\
%     \InfoGANB   & 0.1748    & 0.1762 \\
%     \InfoGANC   & 0.0507    & 0.0655 \\
%     \bottomrule
% \end{tabular}
% }

\begin{table}
	\centering
	\begin{tabular}{lcccl@{\ }l}
		\toprule
% 					& \# Categorical	& \# Continuous	& \# Binary	& Dataset	\\
					& \# Cat.	& \# Cont.	& \# Bin.	& \multicolumn{2}{l}{Dataset}	\\
		\midrule
		\InfoGANA	& 10		& 2			& 0			& \PlainData:	& plain	only \\
		\InfoGANB	& 10		& 3			& 0			& \GlobalData:	& plain + thinning + thickening	\\
		\InfoGANC	& 10		& 2			& 2			& \LocalData:	& plain + swelling + fractures	\\
		\bottomrule
	\end{tabular}
	\caption{Settings for InfoGAN disentanglement experiments}
    \label{tab:infogan_settings}
\end{table}

\subsubsection{Inferential Disentanglement}\label{sec:disentanglement_inferential}
To illustrate how this methodology can be applied in practice to assess disentanglement, we consider two settings. The first is the same as in the MNIST experiment from \citet{Chen2016}, with a 10-way categorical and two continuous latent codes, trained and evaluated on the plain MNIST digits, to which we will refer as \InfoGANA.

The second setting was designed to investigate whether the model could disentangle the concept of \emph{thickness}, by including an additional continuous latent code and training on a dataset with exaggerated thickness variations. We constructed this dataset by randomly interleaving plain, thinned and thickened digit images in equal proportions. Since the perturbations were applied completely at random, we expect a trained generative model to identify that thickness should be largely independent of the other morphological attributes. We refer to this set-up as \InfoGANB. \Cref{tab:infogan_settings} summarises the different experimental settings, for reference.

\begin{figure}
	\centering
	\newcommand{\figscale}{.68}%{.62}
	\begin{subfigure}[t]{.48\linewidth}%{.46\linewidth}
		\centering
		\includegraphics[scale=\figscale]{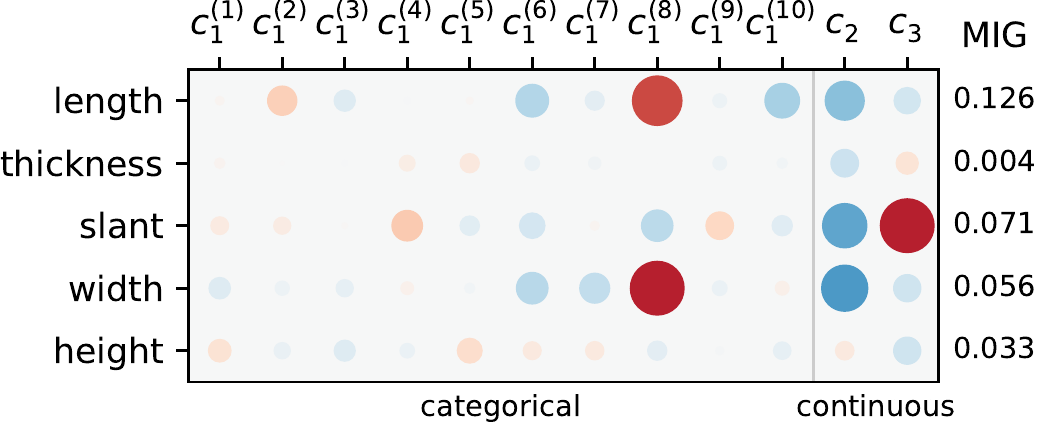}
		\caption{\InfoGANA\ (overall MIG $=$ 0.0582)}
		\label{fig:pcorr_plain_test}
	\end{subfigure} \hfill
	\begin{subfigure}[t]{.5\linewidth}
		\centering
		\includegraphics[scale=\figscale]{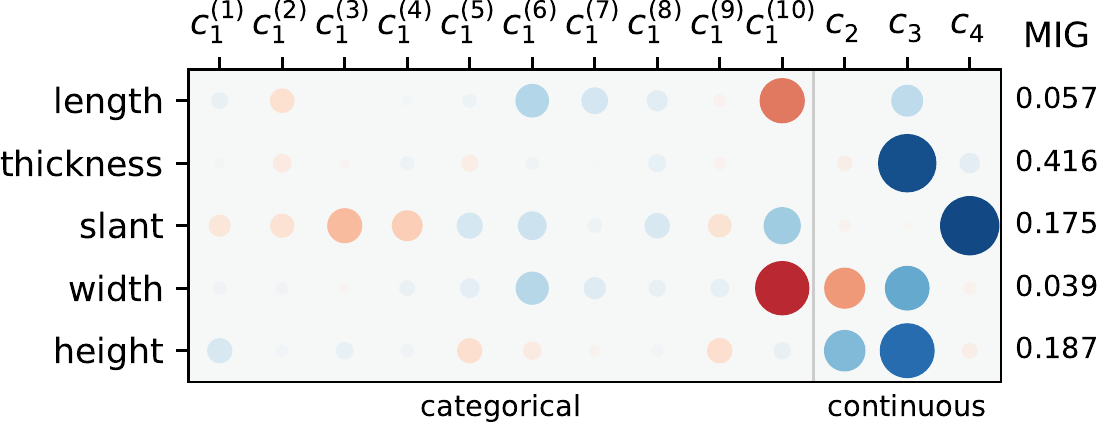}
		\caption{\InfoGANB\ (overall MIG $=$ 0.1748)}
		\label{fig:pcorr_plain_thin_thic_test}
	\end{subfigure}
	\caption{Partial correlations between inferred latent codes and morphometrics of test images. Circle area and colour strength are proportional to correlation magnitude, blue is positive and red is negative. On the right, we indicate the mutual information gap score (MIG) for each attribute.}
	\label{fig:pcorr_test}
\end{figure}

In \cref{fig:pcorr_plain_test}, we see that \InfoGANA{} learned to encode slant mostly in $c_3$, while $c_1^{(8)}$ clearly relates to the `1' class (much narrower digit shape and shorter pen stroke; cf.\ \cref{fig:morpho_distribution}). \Cref{fig:pcorr_plain_thin_thic_test} quantitatively confirms the hypothesis that \InfoGANB's recognition network would learn to separate slant and thickness (in $c_4$ and $c_3$, resp.), the most prominent factors of style variation in this dataset. Interestingly, it shows that $c_3$ also associates with height, as thicker digits also tend to be taller.

Moreover, we observe that the MIG for each attribute correlates with the qualitative interpretation of these partial correlation tables. For example, thickness is represented considerably more compactly in \InfoGANB\ (mainly by $c_3$) than in \InfoGANA\ (highly entangled), exhibiting a correspondingly much higher MIG score.
For \InfoGANB, a noticeable portion of the slant variation was additionally captured by $c_1$, which is reflected in its lower MIG than that for thickness. Lastly, the difference in overall MIG between models validates our qualitative judgement about global disentanglement.
% Note that the MIG score considers the categorical variable, $c_1$, as a single entity, rather than as a collection of binary variables.

\subsubsection{Generative Disentanglement}\label{sec:disentanglement_generative}
The evaluation methodology described above is useful for investigating the behaviour of the \emph{inference} direction of a model, and can readily be used with datasets which include ground-truth generative factor annotations. On the other hand, unless we trust that the inference approximation is highly accurate, this tells us little about the \emph{generative} expressiveness of the model. This is where computed metrics truly show their potential: one can measure generated samples, and investigate how their attributes relate to the latent variables used to create them.

\begin{figure}
	\centering
	\newcommand{\figscale}{.68}%{.62}
	\begin{subfigure}[t]{.48\linewidth}%{.46\linewidth}
		\centering
		\includegraphics[scale=\figscale]{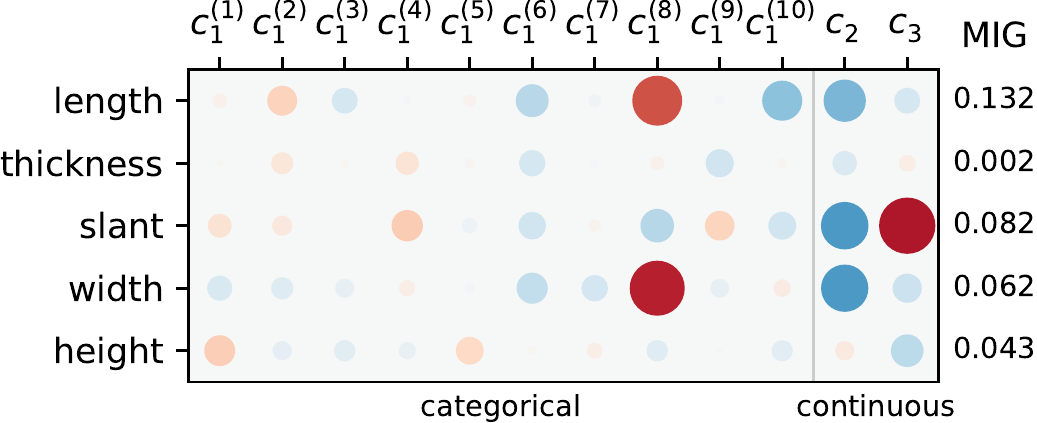}
		\caption{\InfoGANA\ (overall MIG $=$ 0.0641)}
%		 \caption{10-way categorical + 2 continuous codes, plain dataset}
		\label{fig:pcorr_plain_sample}
	\end{subfigure} \hfill
	\begin{subfigure}[t]{.5\linewidth}
		\centering
		\includegraphics[scale=\figscale]{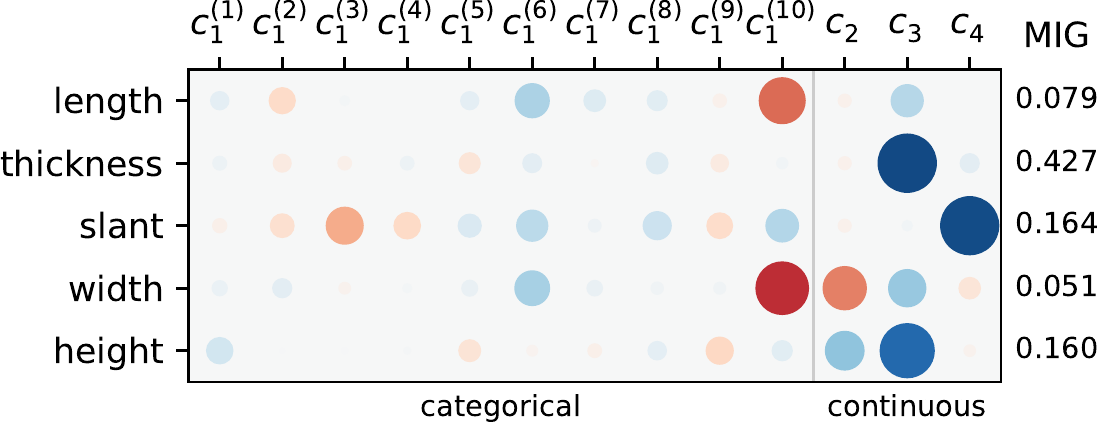}
		\caption{\InfoGANB\ (overall MIG $=$ 0.1762)}
%		 \caption{10-way categorical + 3 continuous codes, plain + thinned + thickened dataset}
		\label{fig:pcorr_plain_thin_thic_sample}
	\end{subfigure}
	\caption{Partial correlations between 1000 sampled latent codes and morphometrics of the corresponding generated images}
	\label{fig:pcorr_sample}
\end{figure}

\Cref{fig:pcorr_sample} shows results for a similar analysis to \cref{fig:pcorr_test}, but now evaluated on \emph{samples} from that model. As the tables are mostly indistinguishable, we may argue that in this case the inference and generator networks have learned to consistently encode and decode the digit shape attributes. Further, the per-attribute and overall MIG scores agree once more with the observed disentanglement patterns.

\begin{figure}[tb]
	\centering
	\newcommand{\figscale}{.39}
	\begin{subfigure}[c]{.52\linewidth}
		\centering
		\includegraphics[scale=\figscale]{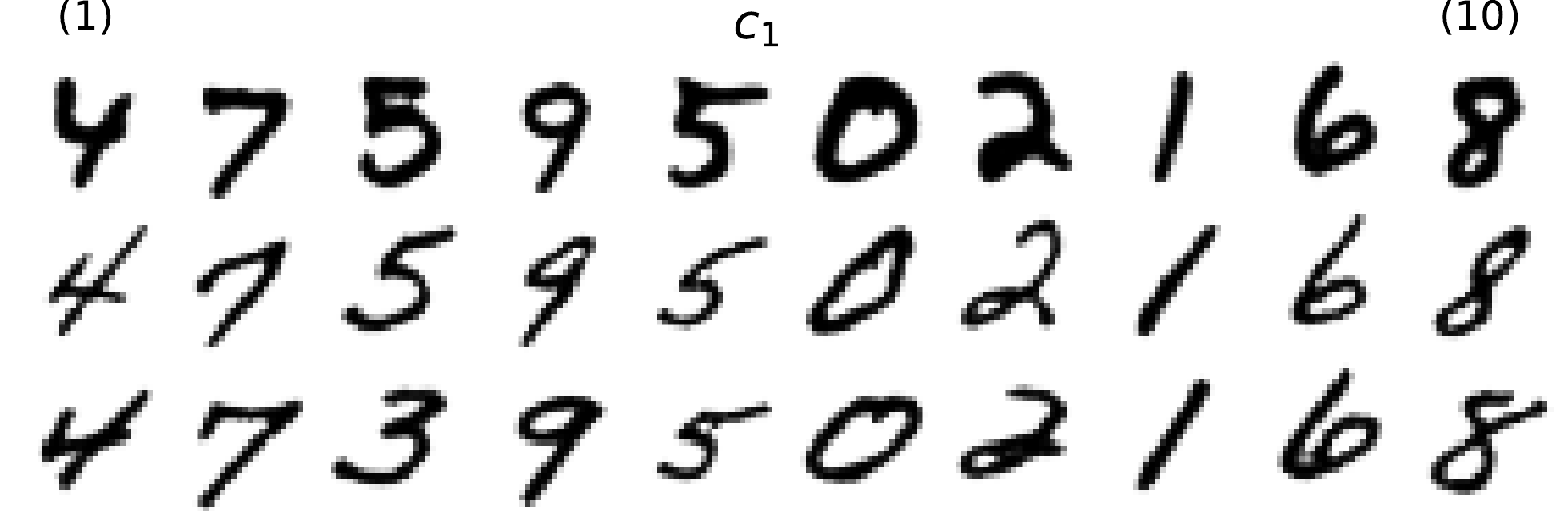} \\[2ex]
		\includegraphics[scale=\figscale]{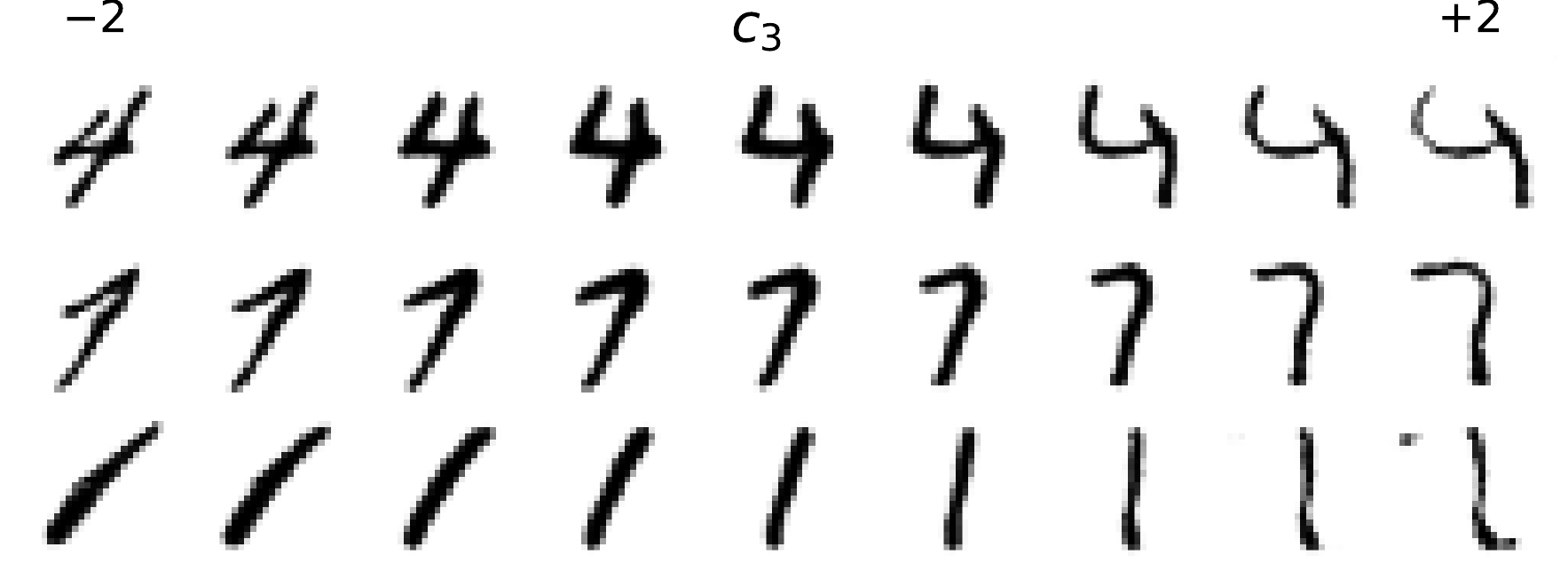}
		\caption{\InfoGANA: one-dimensional traversals of $c_1$ (\emph{top}, `digit type') and $c_3$ (\emph{bottom}, `slant'). Samples in each row share the values of remaining latent variables and noise.}
		\label{fig:traversal_plain_sample}
	\end{subfigure} \hfill
	\begin{subfigure}[c]{.43\linewidth}
		\centering
		\includegraphics[scale=\figscale]{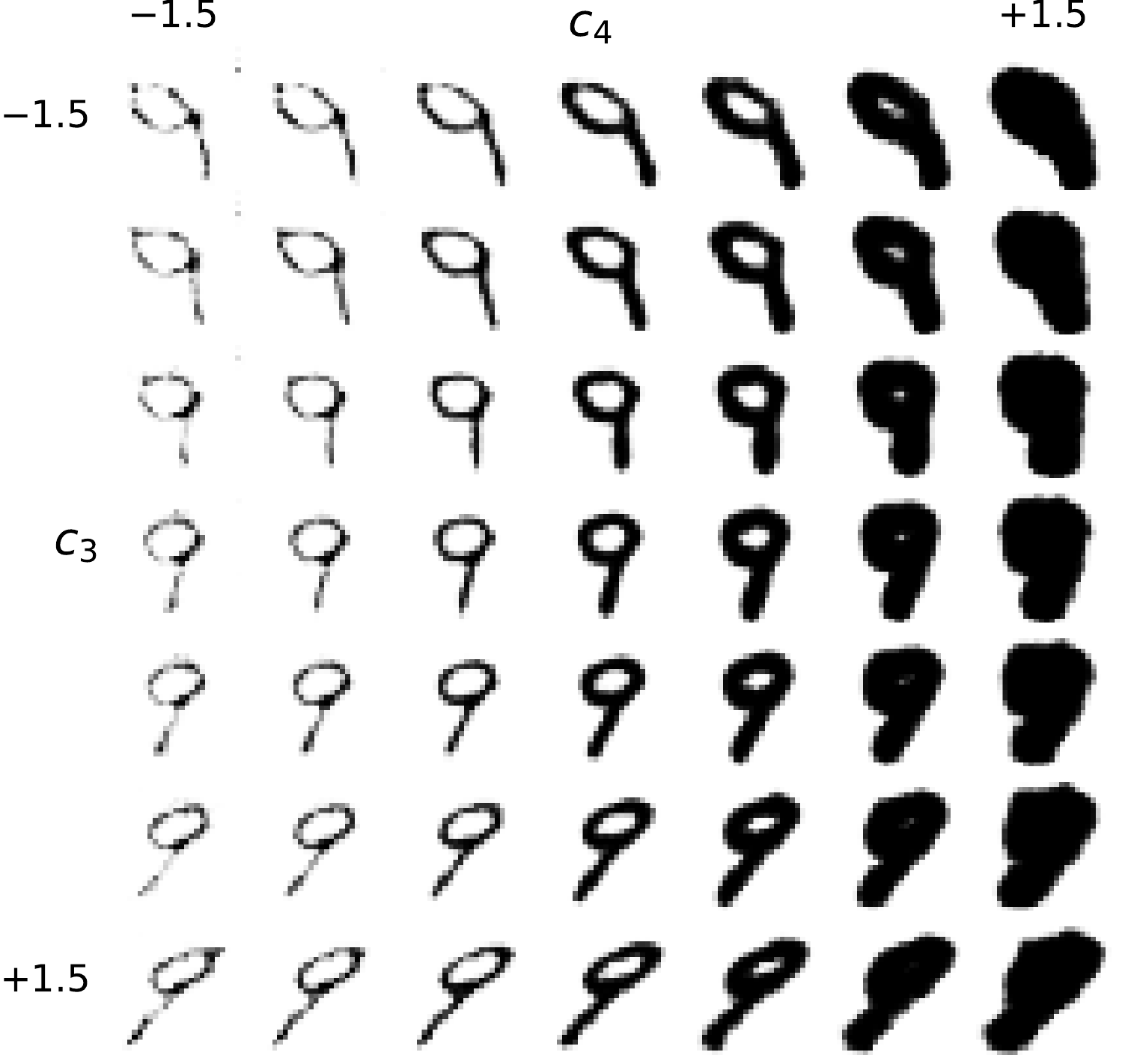}
		\caption{\InfoGANB: two-dimensional traversal of $c_4 \times c_3$ (`thickness' $\times$ `slant')}
		\label{fig:traversal_plain_thin_thic_sample}
	\end{subfigure}
	\caption{InfoGAN latent space traversals}
	\label{fig:traversal_sample}
\end{figure}

As further illustration, \cref{fig:traversal_sample} displays traversals of the latent space, obtained by varying a subset of the latent variables while holding the remaining ones (including noise) constant. With these examples, we can qualitatively verify the quantitative results in \cref{fig:pcorr_sample}. Note that, until now, visual inspection was typically the only means of evaluating disentanglement and expressiveness of the generative direction of image models \citep{Kim2018}.

\subsubsection{Diagnosing Failure}
We also attempted to detect whether an InfoGAN had learned to discover that local perturbations (swelling and fractures) correspond to independent generative factors. To this end, we extended the model formulation with additional Bernoulli latent codes, which may learn to encode presence/absence of each type of local perturbation. The model investigated here, dubbed \InfoGANC\ (cf.\ \cref{tab:infogan_settings}), had a 10-way categorical, two continuous and two binary codes, and was trained with a dataset of plain, swollen and fractured digits (randomly mixed as above).

\begin{figure}[tb]
	\centering
	\newcommand{\figscale}{.66}%{.6}
	\includegraphics[scale=\figscale]{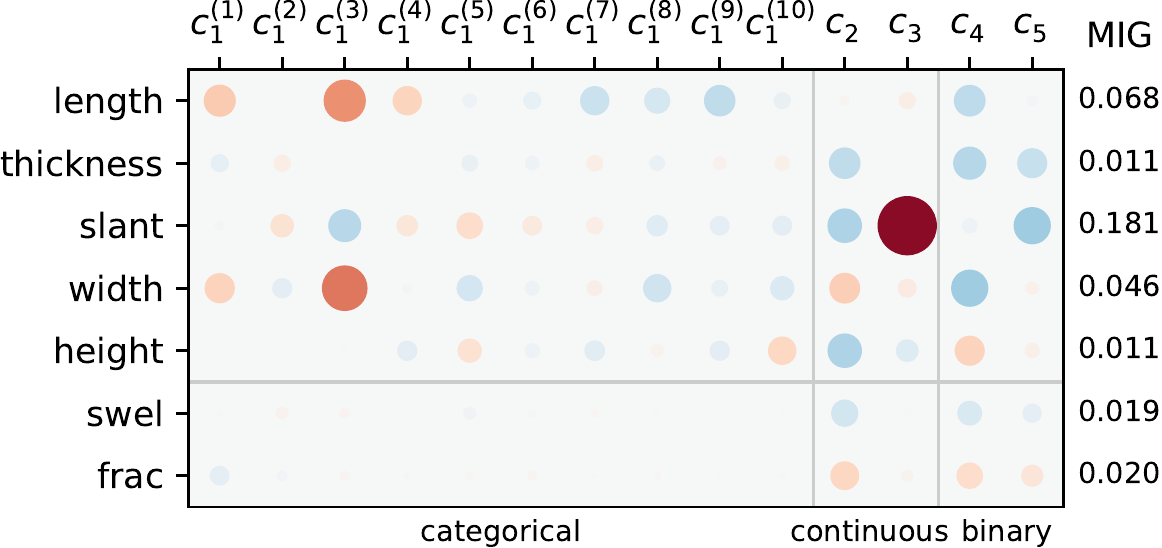}
	\caption{Partial correlations between inferred latent codes and morphometrics of test images for \InfoGANC\ (overall MIG $=$ 0.0507). `\textsf{swel}' and `\textsf{frac}' refer to the binary perturbation labels.}
	\label{fig:pcorr_plain_swel_frac_test}
\end{figure}

Again via inferential partial correlation analysis---now including ground-truth perturbation labels---we can quantitatively verify that this particular model instance was unable to meaningfully disentangle the perturbations (\cref{fig:pcorr_plain_swel_frac_test}, bottom-right block). This is also confirmed by the low MIG for the corresponding rows. In fact, it appears that the addition of the binary variables did not lead to more expressive representations in this case, even impairing the disentanglement of the categorical variables, if compared to \cref{fig:pcorr_plain_test,fig:pcorr_plain_thin_thic_test}, for example.

\subsection{Predictive Model Evaluation}\label{sec:supervised}

% Although the driving motivation for introducing Morpho-MNIST has been the lack of means for quantitative evaluation of generative models, the proposed framework may also be a valuable resource in the context of supervised learning. We conducted several experiments to demonstrate potential applications of these datasets with increased difficulty due to the injected perturbations: standard digit recognition, supervised abnormality detection, and thickness regression. Note such experiments can later serve as baselines for unsupervised tasks such as outlier detection and domain adaptation.

Although the driving motivation for introducing Morpho-MNIST has been the lack of means for quantitative evaluation of generative models, the proposed framework may also be a valuable resource in the context of predictive models. In this section we present several case studies to demonstrate how the proposed perturbations can be useful for systematically evaluating predictive robustness under controlled domain shift and how morphometry enables an entirely new variety of experimental settings and nuanced analyses based on shape stratification. Further we demonstrate how the perturbation labels and morphometrics can serve as prediction targets in their own right for new supervised tasks. For simplicity and reproducibility, we use the same datasets as in the disentanglement experiments (\cref{sec:disentanglement}): plain digits (\PlainData), plain mixed with thinned and thickened digits (\GlobalData), and plain mixed with swollen and fractured digits (\LocalData).

\subsubsection{Robustness}\label{sec:robustness}

% \todo[inline]{Intro paragraph: describe problem setting, 

% `Digits with local perturbations are still clearly identified by humans (see \cref{fig:examples_swel,fig:examples_frac}), but how do different types of classifiers react to these changes?'}

Synthetic perturbations enable fabrication of domain shift scenarios, aiming to systematically evaluate the robustness of predictive models to specific types of variation. For example, digits with local perturbations (swelling and fractures) are still clearly identifiable by humans (see \cref{fig:examples_swel,fig:examples_frac}). However, a practitioner might be interested in investigating how different types of classifiers react to these changes.

Here we compare simple representative methods from four classification paradigms: neighbourhood-based, kernel-based, fully connected, and convolutional. Specifically, we evaluated $k$-nearest-neighbours ($k$NN) using $k=5$ neighbours and $\ell_1$ distance weighting, a support vector machine (SVM) with polynomial kernel and penalty parameter $C=100$, a multi-layer perceptron (MLP) with 784--200--200--$L$ architecture ($L$: number of outputs), and a LeNet-5 \citep{LeCun1998} convolutional neural network (CNN).

Each model is trained for digit recognition once on \PlainData{}, then tested separately on both \PlainData{} and \LocalData{} test datasets, to investigate the effect of \emph{domain shift}. All methods suffer a drop in test accuracy on \LocalData{} (\cref{tab:supervised_results}), and, considering that one third of the images in \LocalData{} are identical to ones in \PlainData{}, the absolute performance drops for the perturbed digits alone are in fact in the order of 1.5\%, 4.9\%, 7.2\%, and 5.4\%, respectively. $k$NN appears to be the most robust to local perturbations, conceivably because they affect only a few pixels, leaving the image distance between neighbours largely unchanged. Contrarily, both the global patterns that SVM and MLP rely on and the CNN's local filter responses may have changed considerably. Such analyses can provide new insights into the behaviour, strengths, and weaknesses of different models, and can be useful to inform new strategies for developing more robust approaches.

\begin{table}[tb]
	\centering
    \newcolumntype{C}{>{\centering\arraybackslash}m{1.2cm}} % centred column with specified width
	\begin{tabular}{lCCCC}
		\toprule
		Test data   & $k$NN & SVM   & MLP   & CNN \\
		\midrule
		\PlainData  & 96.25 & 95.71 & 97.97 & 98.95 \\
		\LocalData  & 95.22 & 92.47 & 93.15 & 95.33 \\
		\bottomrule
	\end{tabular}
% 	\todo[inline]{Should we keep this as a table?}
	\caption{Example results on digit recognition (accuracy, \%) under synthetic domain shift using the proposed data perturbations}
	\label{tab:supervised_results}
\end{table}

% \begin{table}[tb]
% 	\centering
%     \newlength{\taboffset} % multirow vertical offset to compensate cmidrule height
%     \setlength{\taboffset}{0.5\dimexpr\aboverulesep+\belowrulesep+\cmidrulewidth}
%     \newcolumntype{C}[1]{>{\centering\arraybackslash}m{#1}} % centred column with specified width
% 	\begin{tabular}{lC{1.2cm}C{1.2cm}cc}
% 		\toprule
% 				& \multicolumn{2}{c}{Digit Recog.\ (\%)}
% 					& \multirow{2}{*}[-\taboffset]{\shortstack[c]{Abnormality \\ Detection (\%)}}
% 						& \multirow{2}{*}[-\taboffset]{\shortstack[c]{Thickness Reg. \\ (RMSE, pixels)}} \\
% 				\cmidrule(lr){2-3}
% 		Model	& \PlainData & \LocalData \\
% 		\midrule
% 		$k$NN	& 96.25 & 95.22 & 65.10 & 0.4674 \\
% 		SVM		& 95.71 & 92.47 & 77.59 & 0.3647 \\
% 		MLP		& 97.97 & 93.15 & 88.25 & 0.3481 \\
% 		LeNet-5	& 98.95 & 95.33 & 97.53 & 0.2790 \\
% 		\bottomrule
% 	\end{tabular}
% 	\caption{Accuracy on supervised tasks using the proposed data perturbations}
% 	\label{tab:supervised_results}
% \end{table}

\subsubsection{Data Stratification}\label{sec:stratification}

Originally, MNIST could be stratified exclusively by digit class, as this was the only meta-information available. However, with morphometrics, data can be sliced and grouped according to entirely new criteria: we are able to select the shape properties we desire in the training set and also perform finer analyses at test time. It is even possible to combine multiple attributes for multifaceted studies.

To illustrate how these capabilities can be put to use, we present a simplified case study on a hypothetical unsupervised outlier detection scenario. We take the \GlobalData{} dataset, restrict the training data to digits whose thickness is below 3\,px, and evaluate the model's response to test digits across the entire thickness range, especially out-of-distribution samples. Here we assume the true selection criterion is unknown at learning time. Two example frameworks were considered: uncertainty-based and reconstruction-based.

\begin{figure}
    \centering
    \newcommand{\figscale}{.5}
    \setlength{\fboxsep}{0pt}
    \begin{subfigure}[t]{.46\textwidth}
        \centering
        \includegraphics[scale=\figscale]{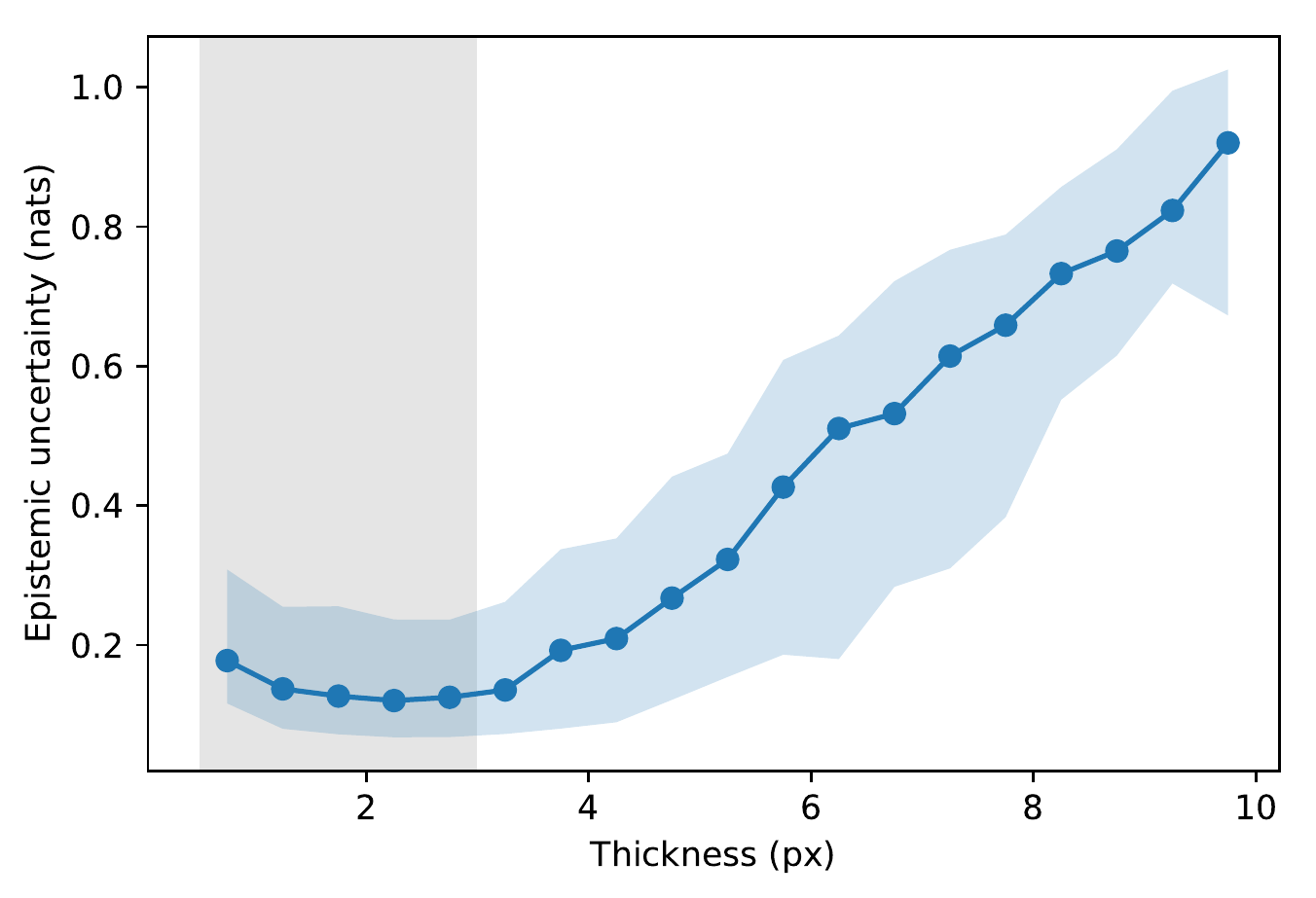}
        \caption{Uncertainty of MC-dropout model}
        \label{fig:strat_mcdropout}
    \end{subfigure}\hfill%
    \begin{subfigure}[t]{.52\textwidth}
        \centering
        \includegraphics[scale=\figscale]{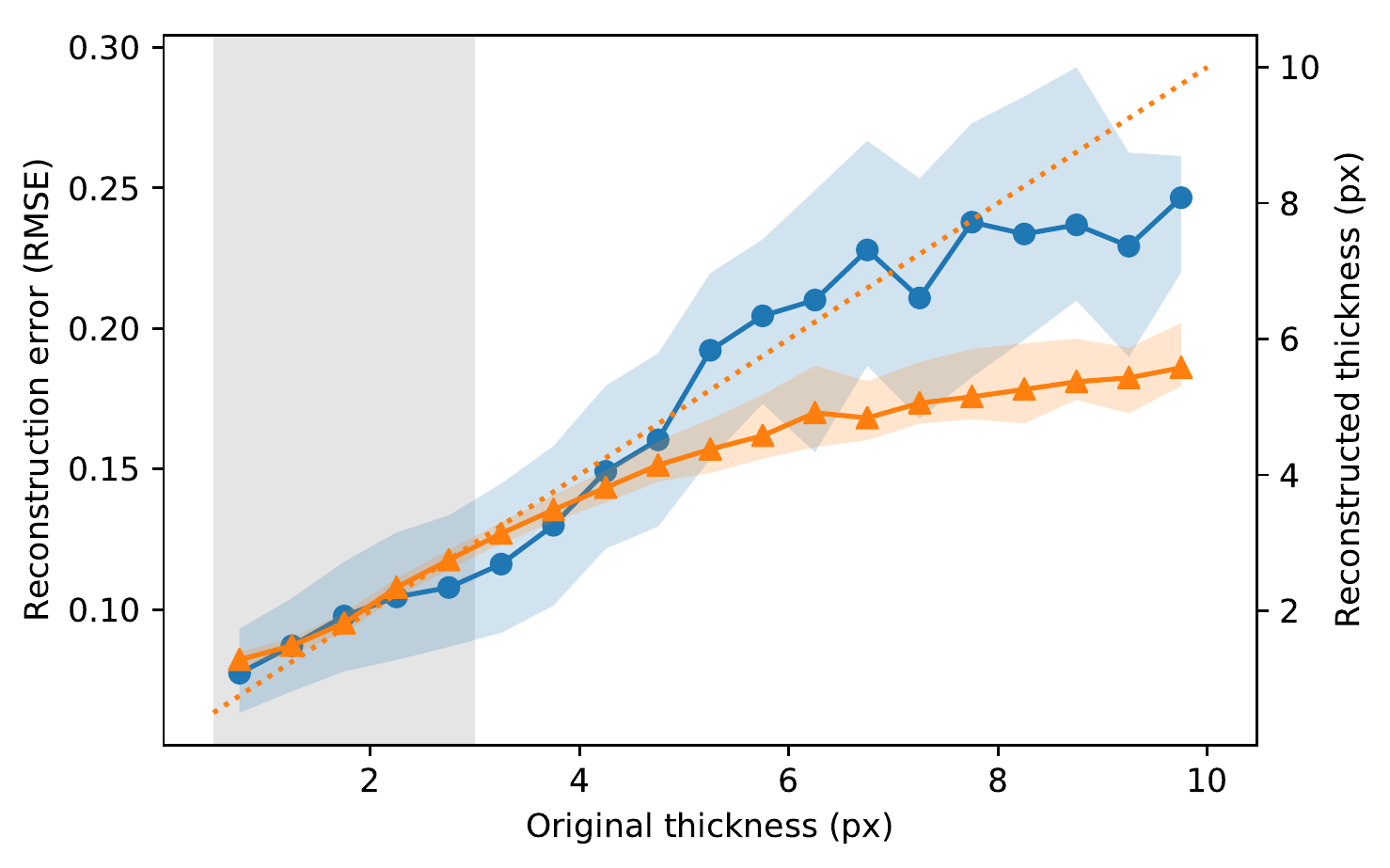}
        \definecolor{mplblue}{HTML}{1f77b4}%
        \definecolor{mplorange}{HTML}{ff7f0e}%
        \caption{Auto-encoder reconstruction error (left, \tikz{\node[fill=mplblue,circle,minimum size=2mm,inner sep=0pt]{};}) and thickness of reconstructed digits (right, \tikz{\fill[mplorange,x=2.2mm,y=2.2mm] (0,0)--(.5,1)--(1,0)--cycle;})}
        \label{fig:strat_recon}
    \end{subfigure}
    \caption{Results for toy outlier detection experiments showcasing data stratification by stroke thickness: training digits were restricted to those thinner than 3\,px (shaded range), and test digits were grouped into 0.5\,px-wide bins for analysis. Lines and shaded regions indicate medians and interquartile ranges.}
\end{figure}

For the first setting, we assume a research question about outlier detection with respect to the digit classification task. We take a LeNet-5 classifier \citep{LeCun1998} and employ Monte Carlo dropout \citep{Kendall2017} to estimate the model's epistemic uncertainty via mutual information \citep{Smith2018}. This metric, which quantifies the model's uncertainty about its parameters when presented with a test input, has been shown to be effective at detecting out-of-distribution samples \citep{Smith2018}. \Cref{fig:strat_mcdropout} shows that the model seems markedly more uncertain on average regarding thicker digits (e.g.\ above 5\,px), although there is no abrupt jump in uncertainty at the inlier boundary. Since we have `ground-truth' outlier labels for the test images based on their thickness, we are also able to evaluate the discriminative power of this score: with an area under ROC curve (AUC) of only 0.67, this approach did not result in a particularly strong outlier detector.

In the second setting, we explored outlier detection with a deep convolutional auto-encoder.%
\footnote{The encoder has three convolutional and one fully connected layers, and the mirrored decoder employs transposed convolutions. The convolutional layers have 3$\times$3 filters with stride 2 and 32--64--128 feature channels, and the bottleneck layer is 10 units wide. %We employ leaky ReLUs and a final logistic activation.
% Leaky ReLU is used for every layer and the final output uses a logistic activation function.
}
It was likewise trained on digits thinner than 3\,px, and we plot in \cref{fig:strat_recon} its reconstruction error for test digits of varying thicknesses. We observe a steeper trend than for the previous model's uncertainty (\cref{fig:strat_mcdropout}), and in fact this auto-encoder's reconstruction error is far more discriminative (AUC $=$ 0.85). In addition, it is also possible to collect morphometrics of the reconstructions and visualise to what extent the original shape attributes are preserved. \Cref{fig:strat_recon} suggests the interesting finding that the reconstructions of thick digits are consistently thinner than the originals, as if the auto-encoder `projected' these digits towards the training subspace to some extent.

% \red{
% Notably, this kind of analysis would not be possible without the availability of \todo{Conclusion paragraph, summarising the take-away message from these two experiments (analysis is only possible because of morphometrics), and generalising to joint stratification by multiple attributes for even more nuanced and task-relevant analyses. Mention potential experiments on inter- and extrapolation, etc.}
% \begin{itemize}
%     \item select training data (e.g.\ experiments on inter- and extrapolation)
%     \item 
% \end{itemize}
% }

\subsubsection{New Prediction Tasks}\label{sec:new_tasks}

% \todo[inline]{Add intro paragraph, move results from Table 3 into text, remove references, and revise thoroughly.}

% \red{As a by-product of the morphometry and perturbation framework, we obtain new covariates that may serve as prediction targets for new supervised tasks. Specifically, we illustrate in this section how one may train models to predict perturbation labels from randomly perturbed images, and how to }

In the previous two sections, we demonstrated how perturbations may be leveraged to systematically evaluate robustness to domain shift and how morphometry can prove valuable for selecting training data and for performing stratified test-time analysis. Here we illustrate how perturbation labels and morphometric attributes can be used for interesting new prediction tasks on MNIST, evaluating the same models from \cref{sec:robustness}.

% 		$k$NN	& 65.10 & 0.4674 \\
% 		SVM		& 77.59 & 0.3647 \\
% 		MLP		& 88.25 & 0.3481 \\
% 		CNN     & 97.53 & 0.2790 \\

As our first example, we define a supervised abnormality detection task, using the \LocalData{} dataset, to predict whether a digit is normal or perturbed (swollen or fractured)---compare with lesion detection in medical scans. The top row of \cref{tab:new_tasks_results} indicates that the CNN was able to detect abnormalities with high accuracy, likely thanks to local invariances of its architecture, whereas the shallow classifiers, $k$NN and SVM, performed remarkably worse. Note how the performance is generally much lower than for digit classification, revealing the higher difficulty of this binary problem compared to the ten-class recognition task.

\begin{table}[tb]
	\centering
    \newcolumntype{C}{>{\centering\arraybackslash}m{1.2cm}} % centred column with specified width
	\begin{tabular}{lCCCC}
		\toprule
		Prediction task             & $k$NN & SVM   & MLP   & CNN \\
		\midrule
		Abnormality detection (\%)  & 65.10 & 77.59 & 88.25 & 97.53 \\
		Thickness regression (RMSE) & 0.467 & 0.365 & 0.348 & 0.279 \\
		\bottomrule
	\end{tabular}
% 	\todo[inline]{Should we keep this as a table?}
	\caption{Example results on some of the new MNIST prediction tasks}
	\label{tab:new_tasks_results}
\end{table}

% The abnormality detection task is, using the \LocalData{} dataset, to predict whether a digit is normal or perturbed (swollen or fractured)---compare with lesion detection in medical scans. \Cref{tab:supervised_results} (third column) indicates that LeNet-5 is able to detect abnormalities with high accuracy, likely thanks to local invariances of its convolutional architecture. Note that all scores (especially the simpler models') are lower than digit classification accuracy, revealing the (possibly surprising) higher difficulty of this binary classification problem compared to the ten-class digit classification.
% Note that this time the perturbations are the prediction targets themselves, rather than nuisance factors. 
% The translation invariance inherent to LeNet-5 may help in identifying these local changes, while SVM and $k$NN methods may have difficulty in cleanly separating these classes.
% The hierarchical integration of spatial information and translation invariance inherent to LeNet-5 may help in cases of tightly closed loops or incomplete loops in original MNIST digits. Meanwhile SVM and KNN methods may mistake these for swelling or fractures respectively.

Finally, we also constructed a regression task for digit thickness using the \GlobalData{} dataset, mimicking medical imaging tasks such as estimating brain age from cortical grey matter maps. Since this is a non-trivial task requiring some awareness of local geometry, it is unsurprising that the convolutional model outperformed the others, which rely on holistic features (\cref{tab:new_tasks_results}, bottom row).

\section{Discussion and Perspectives}\label{sec:discussion}

% \todo[inline]{Paragraph summarising the conclusions from \cref{sec:evaluation}: through concrete worked examples, we demonstrated how morphometry and perturbations can be exploited for qualitative and, more importantly, \emph{quantitative} evaluation of generative models in terms of sample diversity and disentanglement, and for novel experimental settings with predictive models.

% MNIST and similar rasterised shape datasets.}

Through concrete examples, we demonstrated in \cref{sec:evaluation} how morphometry and perturbations can be exploited 1) for qualitative and, more importantly, \emph{quantitative} evaluation of generative models in terms of sample diversity and disentanglement, and 2) in developing novel experimental protocols for analysing predictive models. The presented methodologies are broadly applicable to datasets of rasterised shapes beyond MNIST.

\begin{figure}
    \centering
    \begin{subfigure}[b]{.19\textwidth}
        \centering
        \includegraphics[width=\textwidth, trim={10mm 0 0 0}, clip]{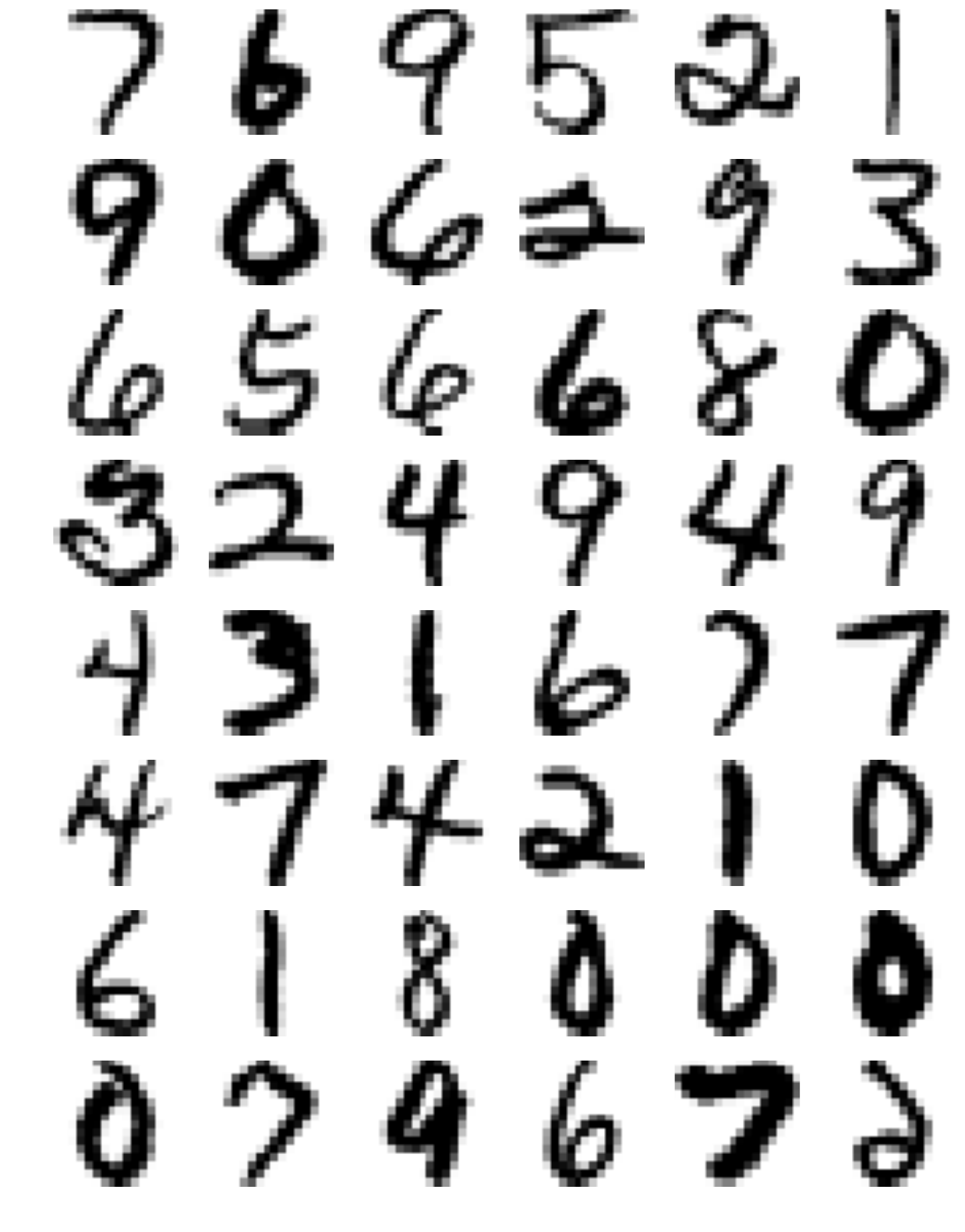}
        \caption{USPS digits}
    \end{subfigure}\hfill%
    \begin{subfigure}[b]{.79\textwidth}
        \centering
        \includegraphics[width=\textwidth]{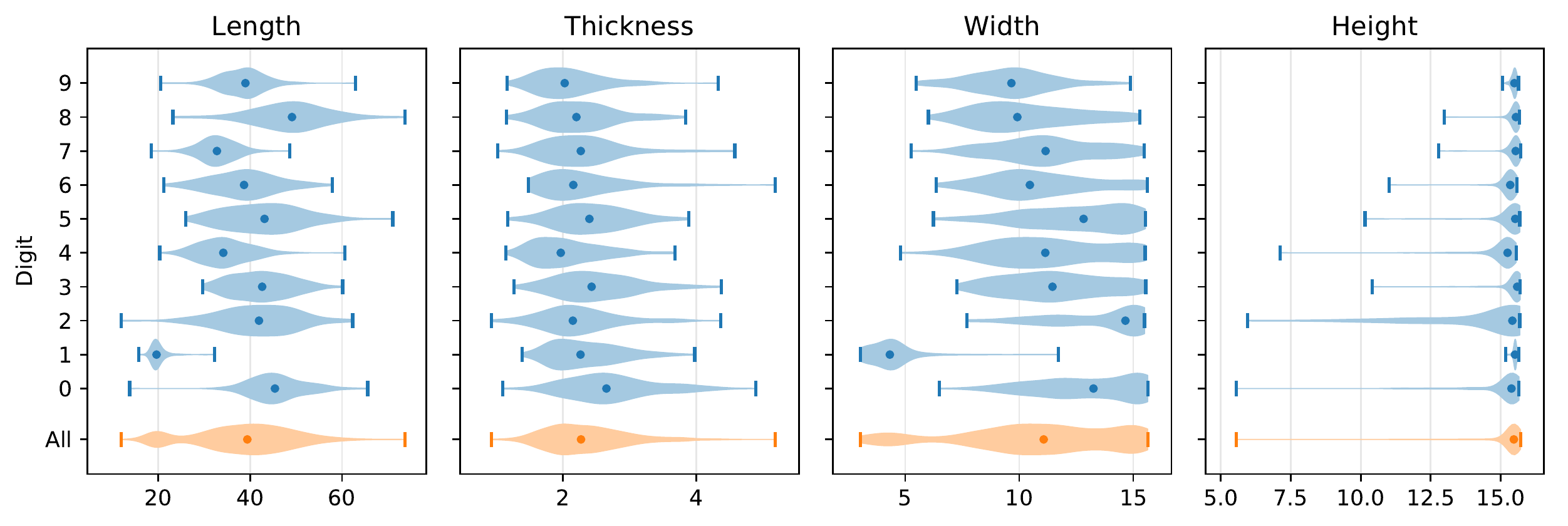}
        \caption{Distribution of morphological attributes for USPS test images}
    \end{subfigure}%
    \caption{Application of the proposed morphometrics to the USPS dataset}
    \label{fig:usps}
\end{figure}

% In addition, we show that our morphometrics are applicable to other datasets beyond MNIST.
For example, we apply our exact same morphometry pipeline to handwritten digits from the USPS dataset \citep{LeCun1990}, also composed of small grayscale images (16$\times$16). Example images from this dataset and distributions of length, thickness, width, and height can be seen in \cref{fig:usps}. Slant was excluded because the USPS digits were already deslanted, with residual variation in the order of $\pm1^\circ$. Another popular dataset to which our morphometrics can be directly applied is Omniglot \citep{Lake2015}, consisting of handwritten symbols from 50 different alphabets.

% We also adapted the morphometric perspective for the dSprites dataset \citep{Matthey2017}, measuring centroid location $(\mi, \mj)$ and scale $\sqrt{\vphantom{s}\smash{\sii\sjj}}$ of the shapes (cf.\ \cref{sec:slant}). We show how our measurements correlate perfectly with the provided ground truth for real data, and speculate\todo{...?}\ that they would be highly robust for generated samples. We conjecture that 

Besides, the general perspective of measuring attributes directly from images is evidently not limited to datasets of handwritten characters. Consider for example the dSprites dataset \citep{Matthey2017}, often used in disentanglement research, which contains 64$\times$64 synthetic binary images of small shapes with ground-truth annotations. Attributes such as location and scale can easily be measured based on image moments (see \cref{sec:slant}). They are expected to be robust even for blurry generated samples. As discussed previously, while the annotations provided with such data facilitate evaluation of inferential disentanglement, these measurements enable the unprecedented analysis of generative behaviour.

An obvious limitation of morphometry is the tacit assumption that generated samples resemble true data closely enough that the shape measurements are indeed meaningful. What happens if the generative model under study is underfitted and produces nonsensical images? In such case, one expects the joint distribution of sample morphometrics to be strikingly different from that of true data, even if some individual metrics may have been extracted successfully (e.g.\ width and height of a reasonably sized blur). Like mode collapse, this issue can be diagnosed straightforwardly---for example with an analysis similar to that from \cref{sec:diversity}---and may constitute an interesting finding by itself.

\section{Conclusion}\label{sec:conclusion}

% With Morpho-MNIST we provide a number of mechanisms to quantitatively assess representation learning with respect to measurable factors of variation in the data. We believe that this \modified{may prove} an important asset for future research on generative models, and we emphasise that the proposed morphometrics can be used \emph{post hoc} to evaluate already trained models, potentially revealing novel insights about their behaviour.

% A similar morphometry approach could be used with other datasets such as dSprites, e.g.\ estimating shape location and size, number of objects/connected components. Perhaps some generic image metrics may be useful for analysis on other datasets, e.g.\ relating to sharpness or colour diversity, or we could even consider using the output of object detectors (analogously to the Inception-based scores; e.g.\ number/class of objects, bounding boxes etc.). Analogous descriptive metrics could also be designed for non-image data.\todo{Move to \cref{sec:discussion}?}

With Morpho-MNIST we provide a number of mechanisms to quantitatively assess representation learning with respect to measurable factors of variation in the data. We believe that this may prove an important asset for future research on generative models, and will hopefully also be useful in evaluating predictive models. We emphasise that the proposed morphometrics can be used \emph{post hoc} to evaluate already trained models, potentially revealing novel insights about their behaviour.
In addition, we have shown that our morphometry and perturbation framework easily extends to datasets other than MNIST. We hope that the case studies presented here may also invite research into developing metrics for other kinds of images beyond rasterised 2D shapes or even to non-imaging data, paving the way to more widespread analysis of true generative performance.

\acks{This work was supported by CAPES, Brazil (BEX 1500/2015-05) and received funding from the European Research Council (ERC) under the European Union's Horizon 2020 research and innovation programme (grant agreement No 757173, project MIRA, ERC-2017-STG).}

\appendix
\numberwithin{figure}{section}
\section{Morphometrics of Plain and Perturbed Datasets}\label{app:pert_morpho}
\vfill
\begin{figure}[h]
	\centering
    \includegraphics[width=\textwidth]{fig/distributions_plain_train.pdf}
    \includegraphics[width=\textwidth]{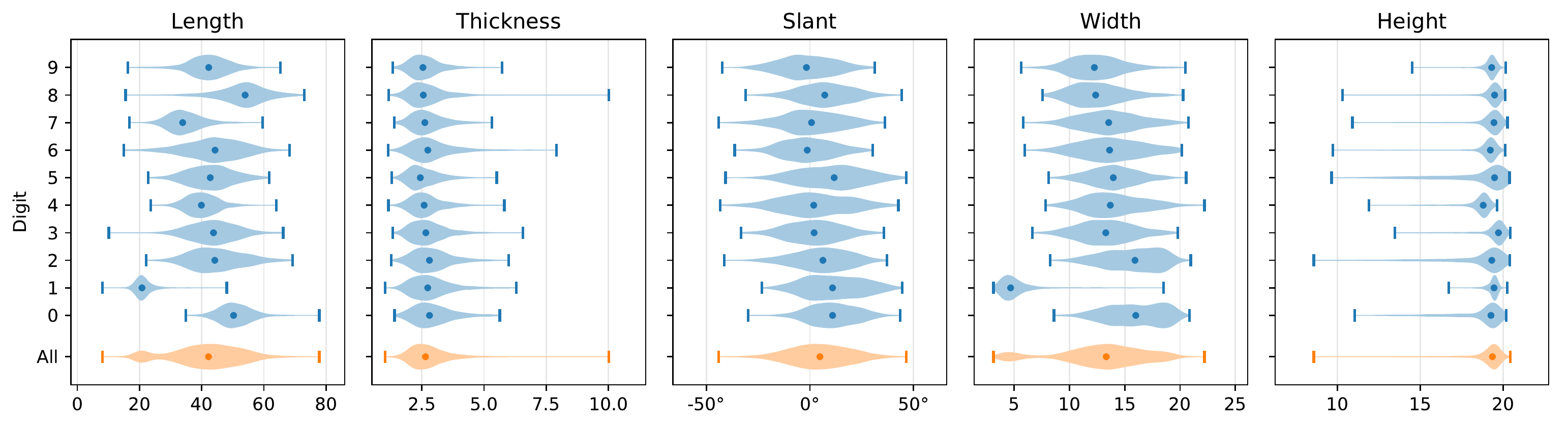}
    \caption{Distribution of morphological attributes for plain MNIST digits. \emph{Top:} training set; \emph{bottom:} test set.}
\end{figure}
\vfill
\begin{figure}[h]
	\centering
    \includegraphics[width=\textwidth]{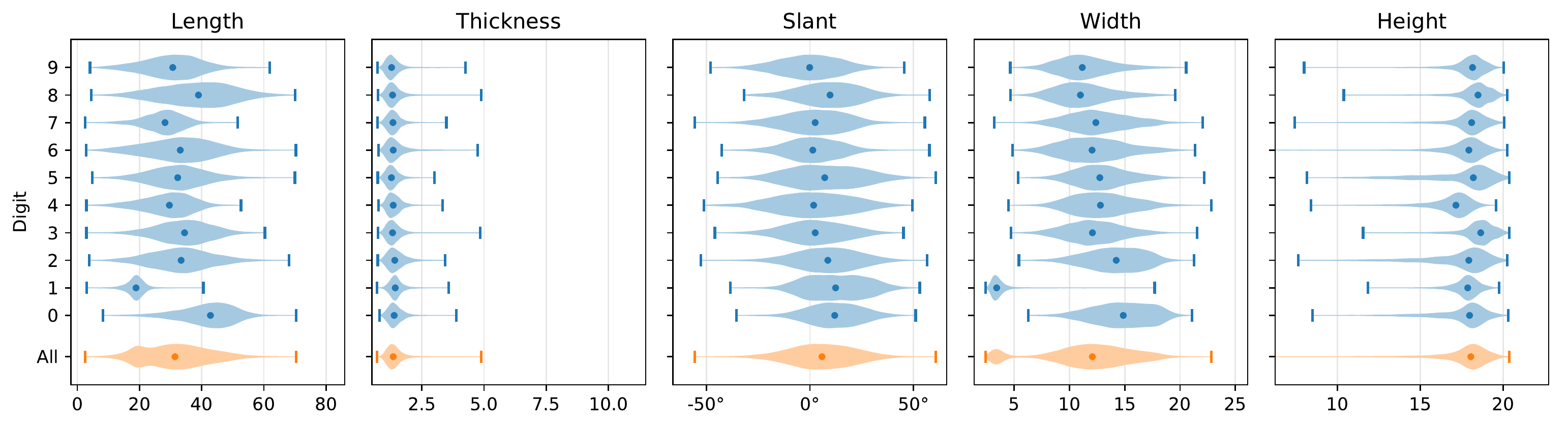}
    \includegraphics[width=\textwidth]{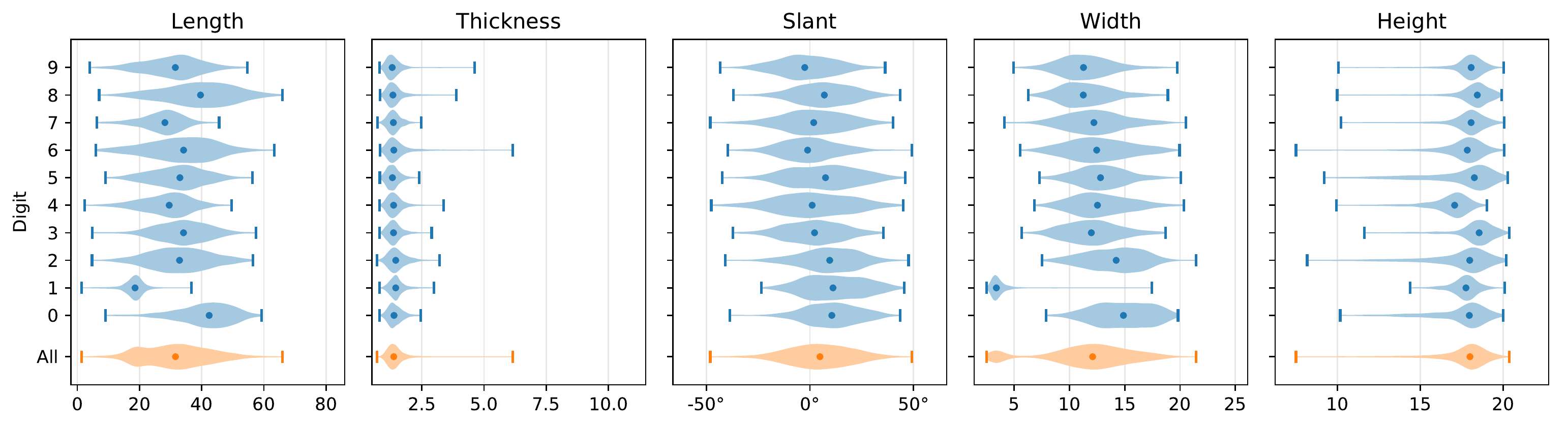}
    \caption{Distribution of morphological attributes for thinned MNIST digits. \emph{Top:} training set; \emph{bottom:} test set.}
\end{figure}

\begin{figure}[h]
	\centering
    \includegraphics[width=\textwidth]{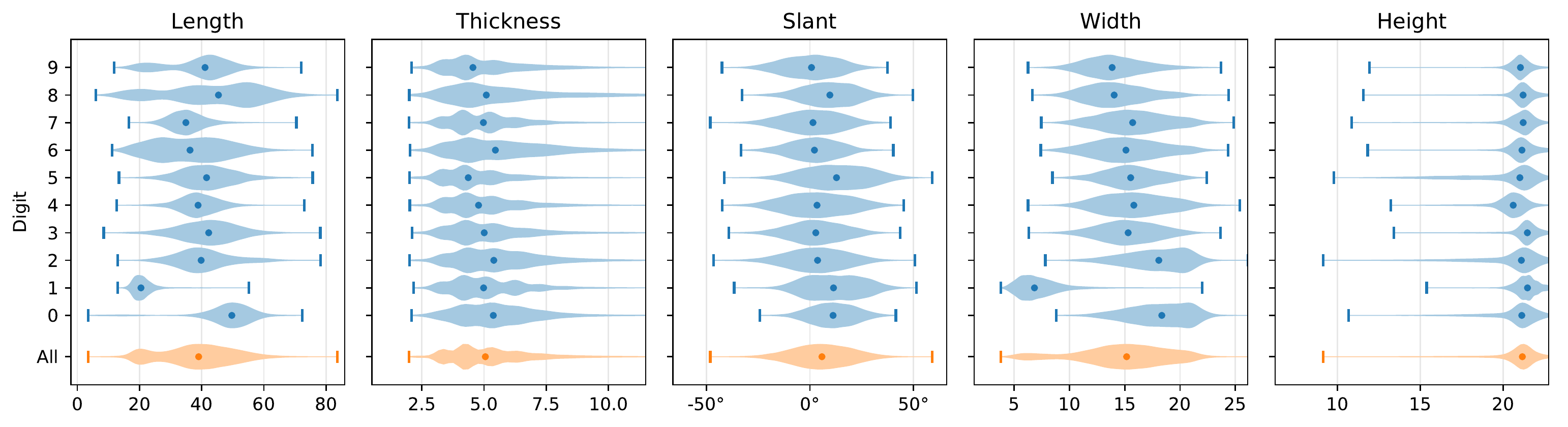}
    \includegraphics[width=\textwidth]{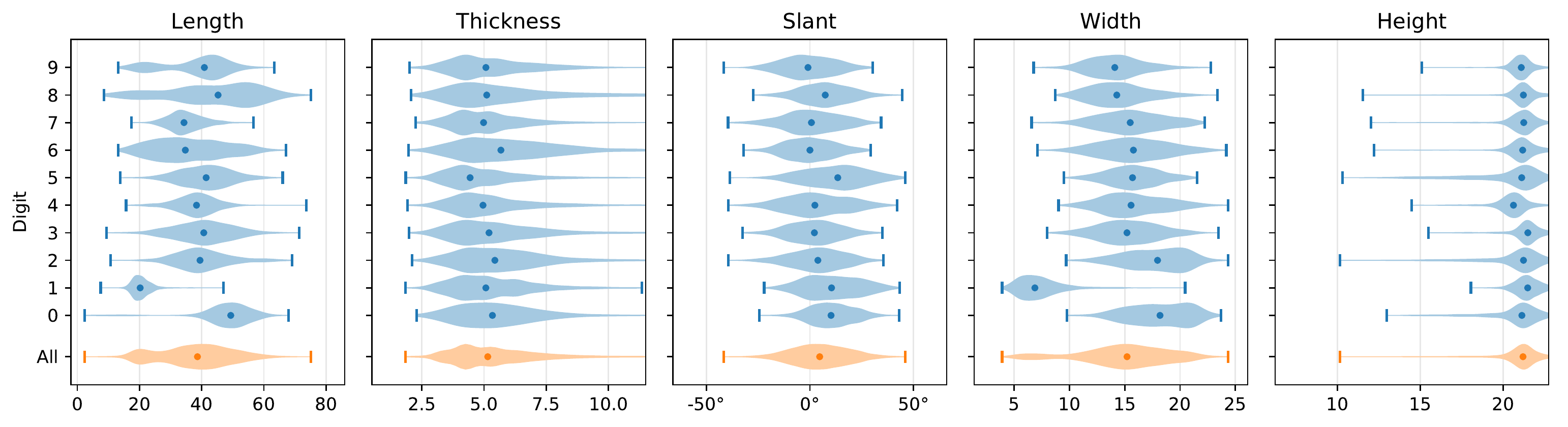}
    \caption{Distribution of morphological attributes for thickened MNIST digits. \emph{Top:} training set; \emph{bottom:} test set.}
\end{figure}

\begin{figure}[h]
	\centering
    \includegraphics[width=\textwidth]{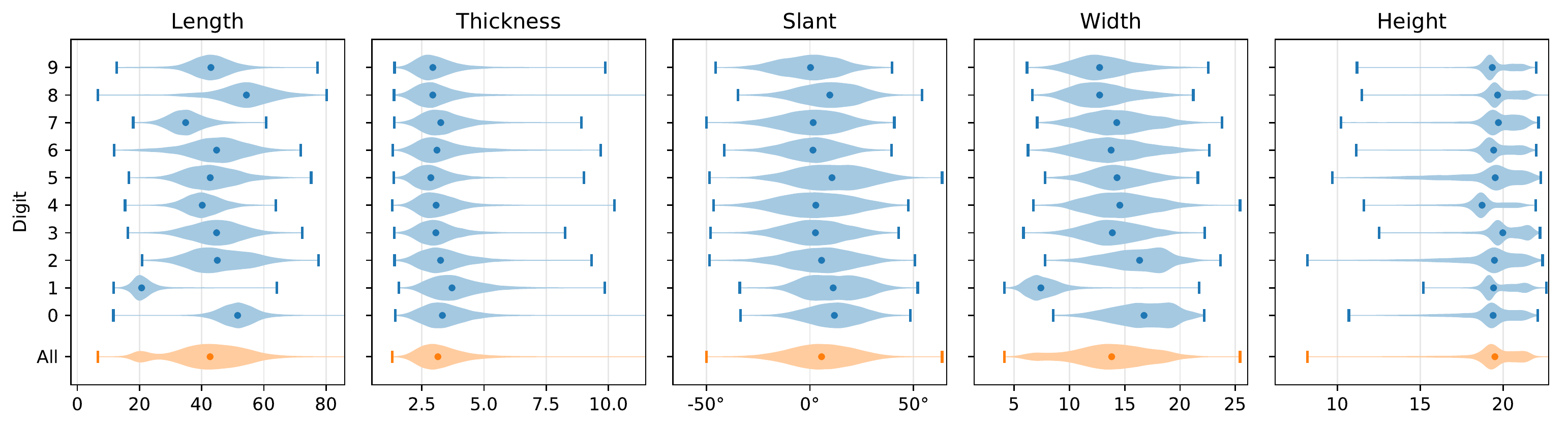}
    \includegraphics[width=\textwidth]{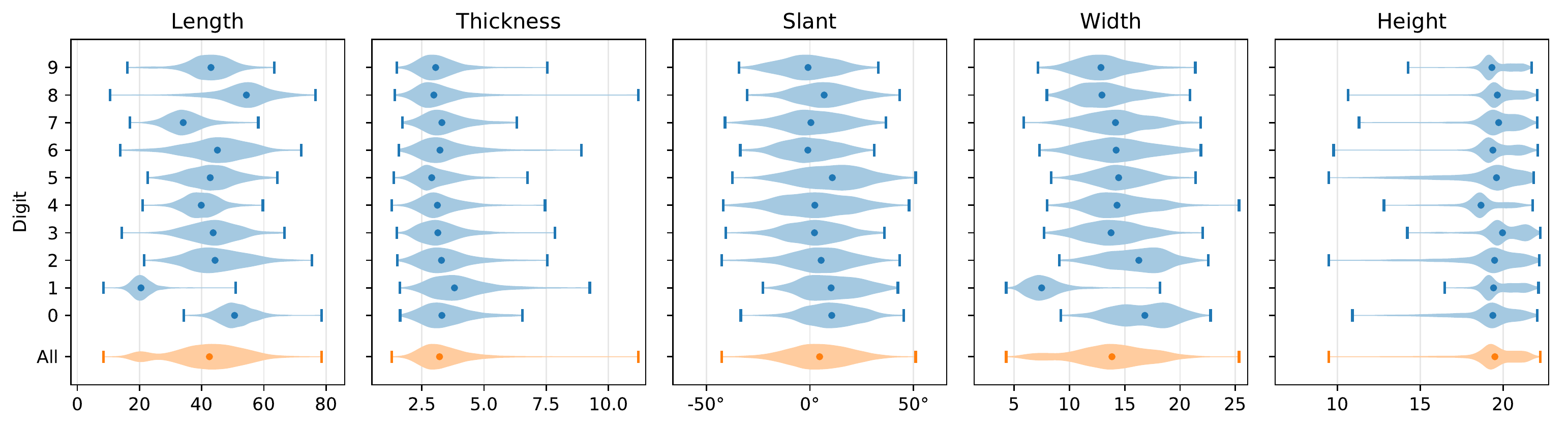}
    \caption{Distribution of morphological attributes for swollen MNIST digits. \emph{Top:} training set; \emph{bottom:} test set.}
\end{figure}

\begin{figure}[h]
	\centering
    \includegraphics[width=\textwidth]{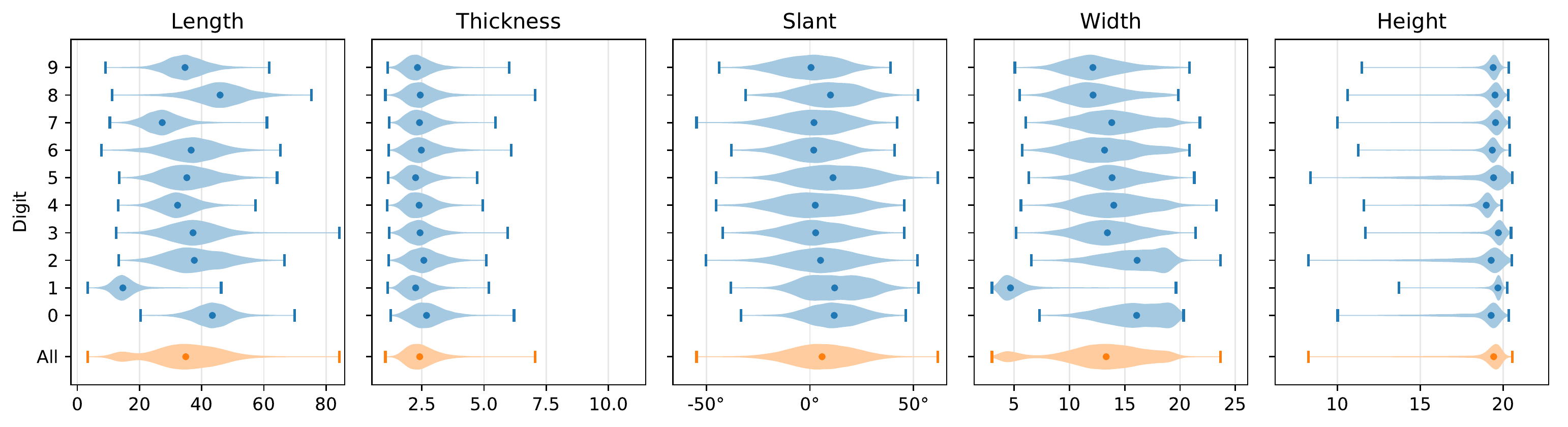}
    \includegraphics[width=\textwidth]{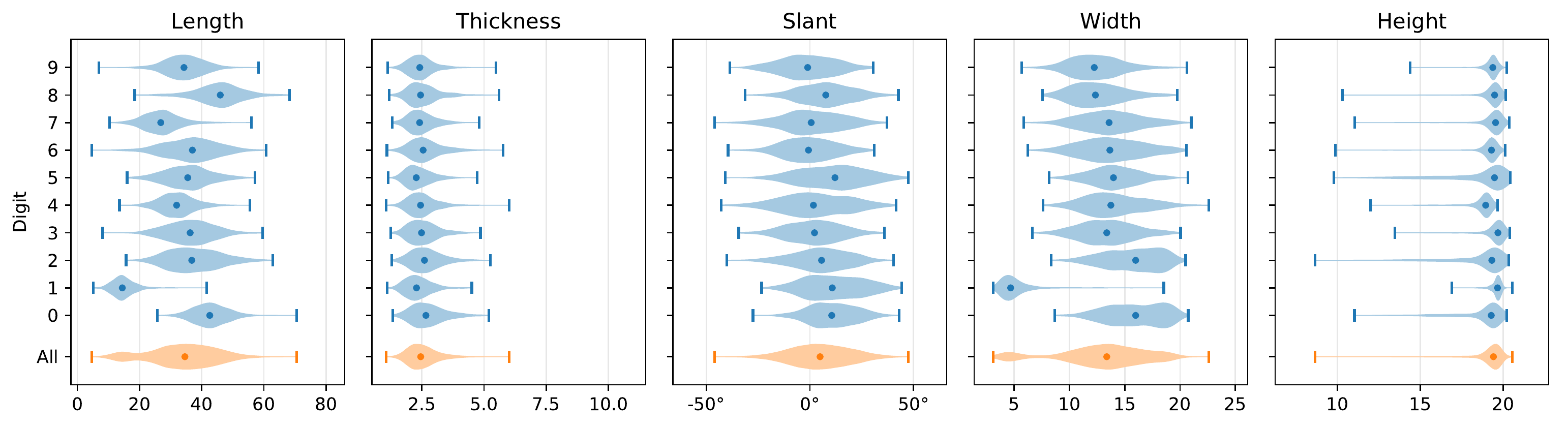}
    \caption{Distribution of morphological attributes for fractured MNIST digits. \emph{Top:} training set; \emph{bottom:} test set.}
\end{figure}
\clearpage

% \clearpage
\section{Perturbation Examples}\label{app:pert_examples}
{\newcommand{\figscale}{.54}
\begin{figure}[h]
	\centering
    \includegraphics[scale=\figscale]{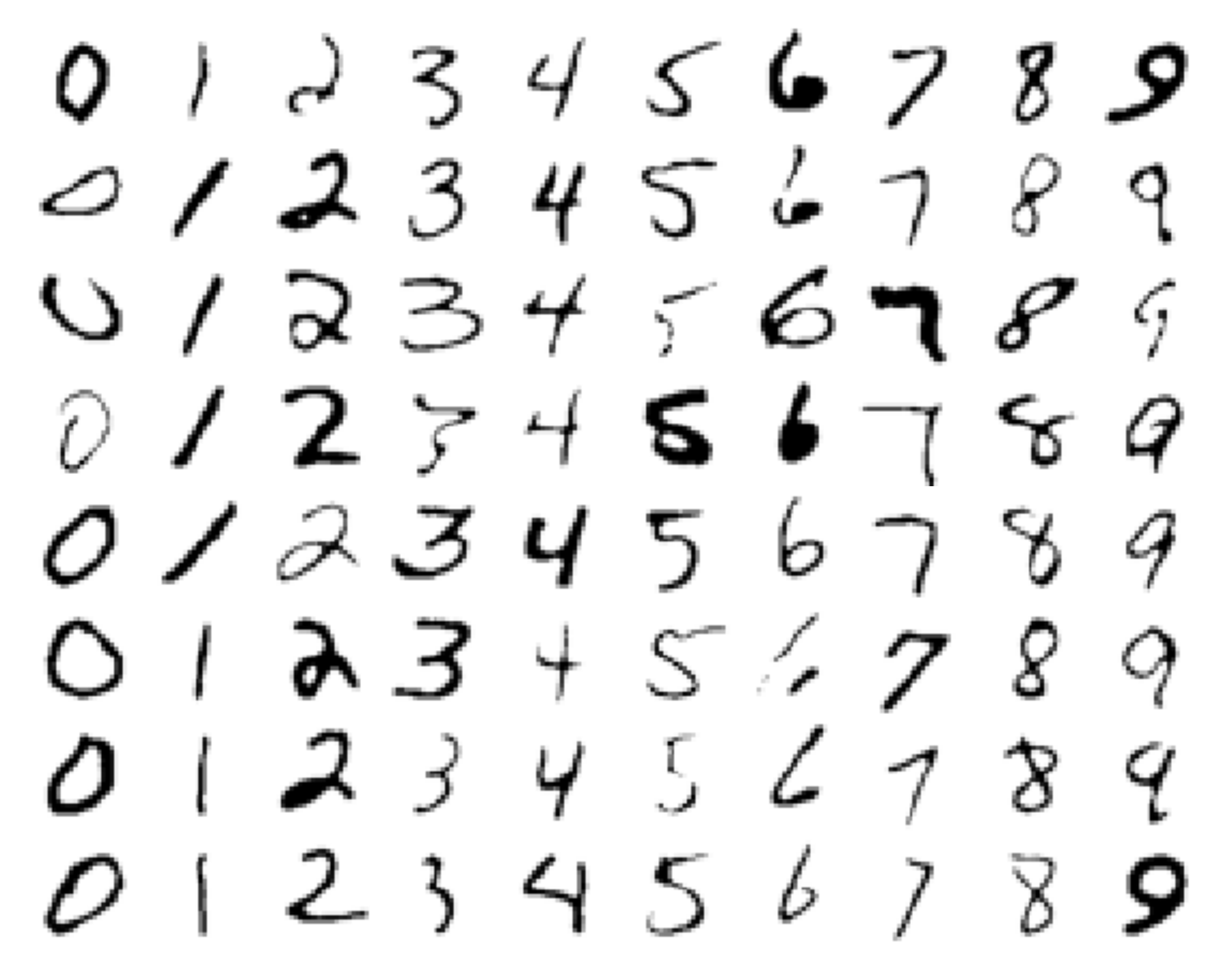}
    \caption{Examples of globally thinned digits}
    \label{fig:examples_thin}
\end{figure}

\begin{figure}[h]
	\centering
    \includegraphics[scale=\figscale]{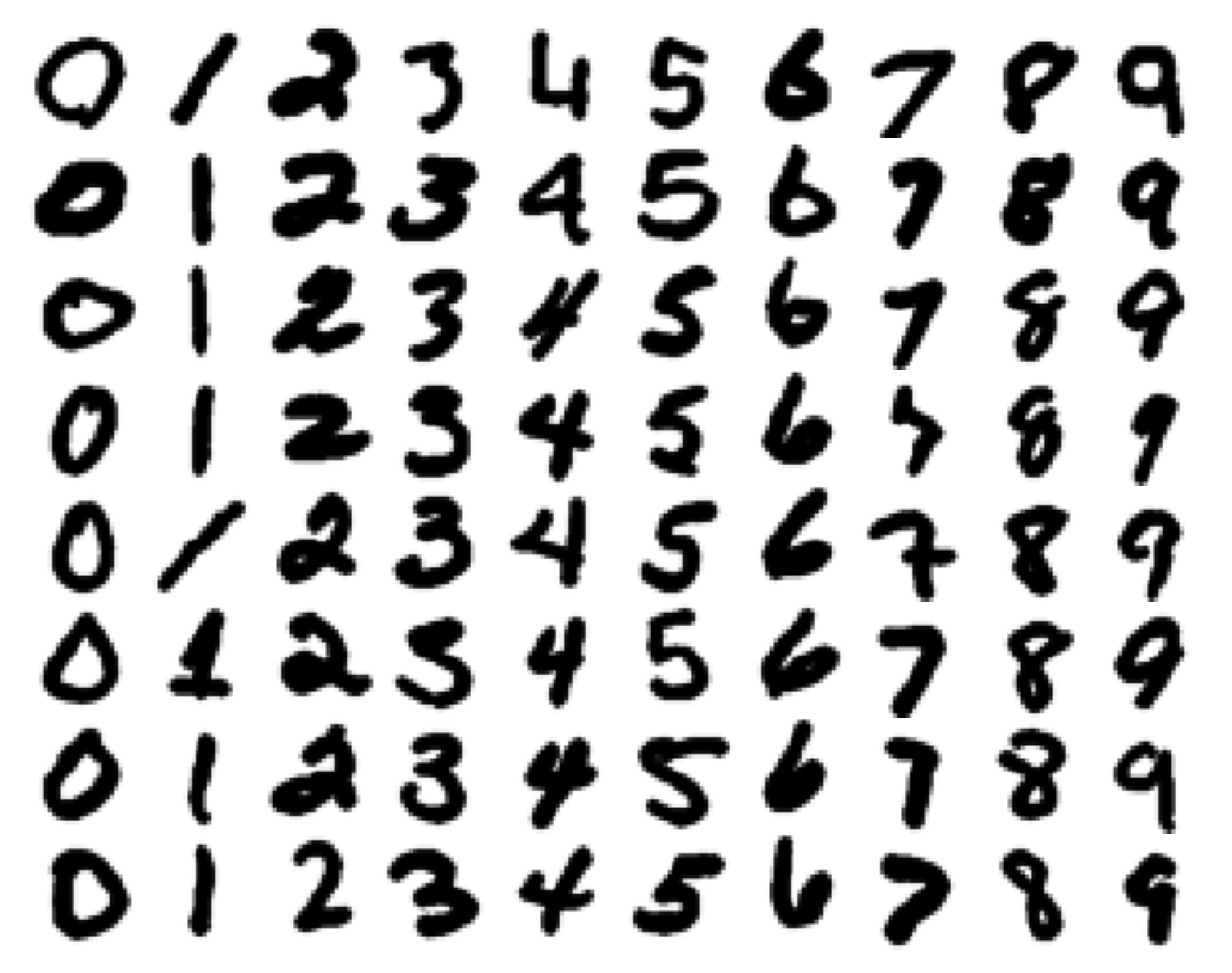}
    \caption{Examples of globally thickened digits}
    \label{fig:examples_thic}
\end{figure}

\begin{figure}[h]
	\centering
    \includegraphics[scale=\figscale]{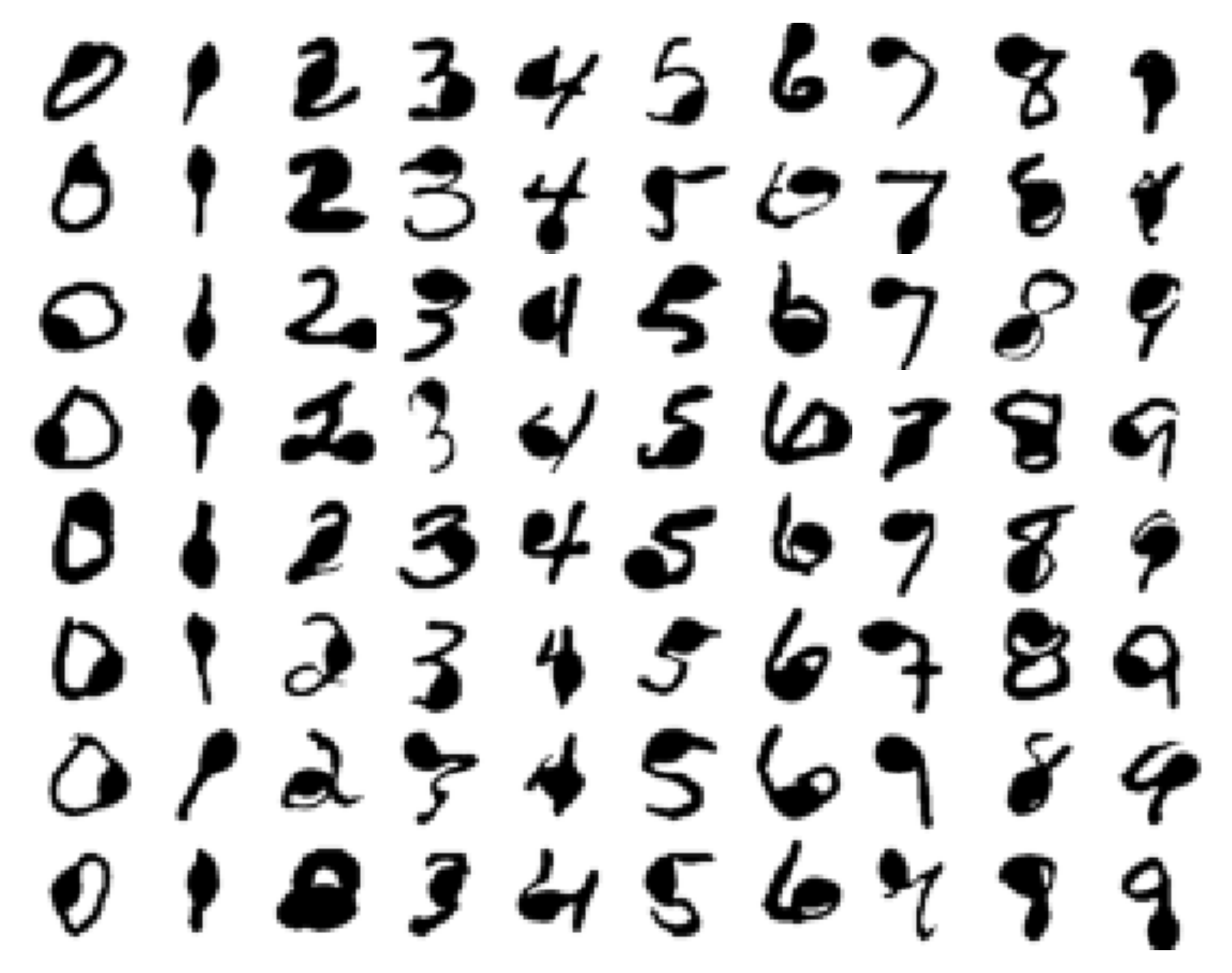}
    \caption{Examples of digits with local swelling}
    \label{fig:examples_swel}
\end{figure}

\begin{figure}[h]
	\centering
    \includegraphics[scale=\figscale]{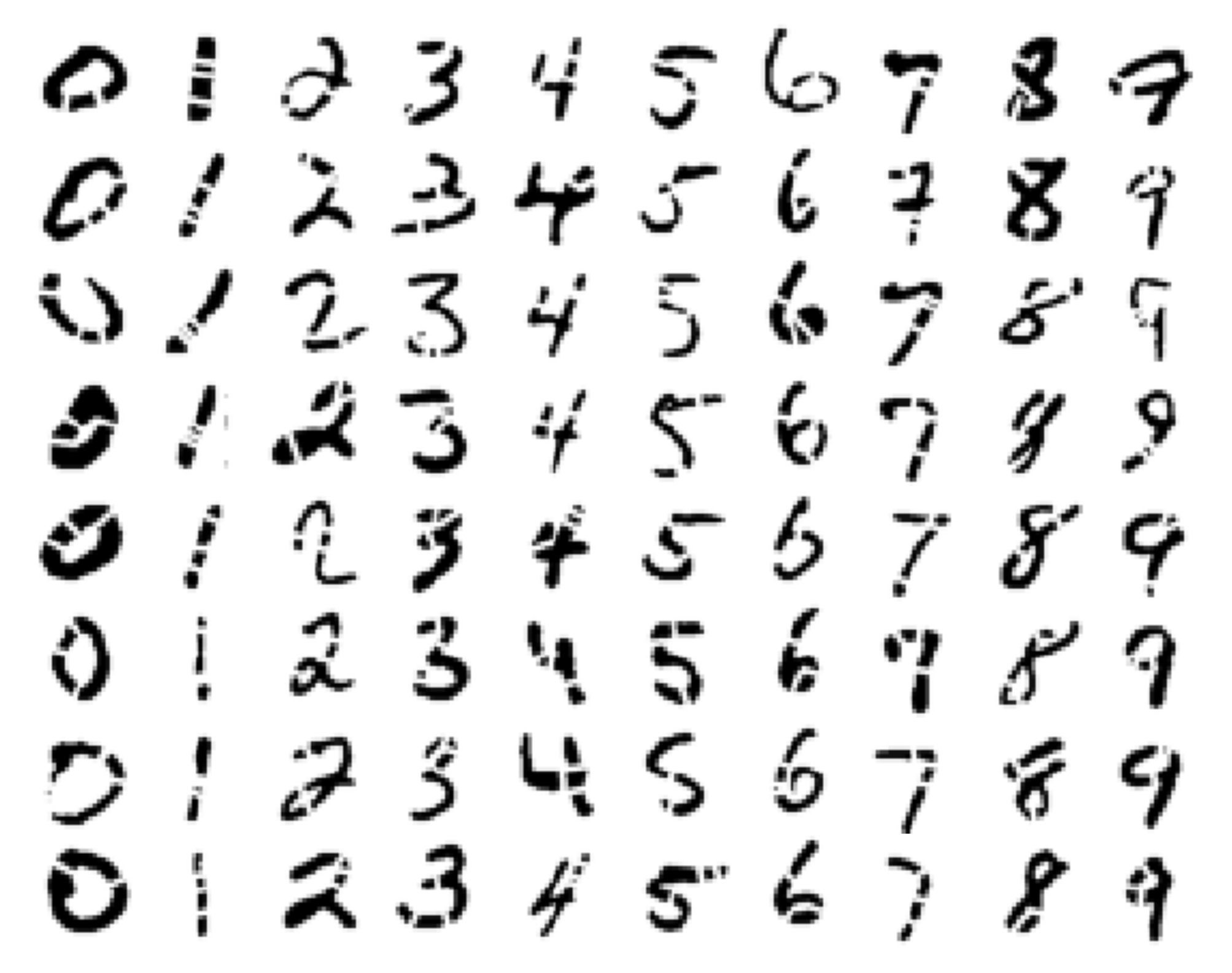}
    \caption{Examples of digits with local fractures}
    \label{fig:examples_frac}
\end{figure}
}
\clearpage

\section{MMD Details}\label{app:kernel}

We employed a Gaussian product kernel with bandwidths derived from Scott's rule, analogously to the KDE plots in \cref{fig:sample_diversity}. Scott's rule of thumb defines the bandwidth for a density estimation kernel as $N^{-1/(D+4)}$ times the standard deviation in each dimension, where $N$ and $D$ denote sample size and number of dimensions \citep[Eq.~6.42]{Scott1992}. We determine the KDE bandwidths separately for real and sample data, then add their squares to obtain the squared bandwidth of the MMD's Gaussian kernel, as it corresponds to the \emph{convolution} of the density estimation kernels chosen for each set of data. See \citet[\S 3.3.1]{Gretton2012} for further details on the relation between MMD and $L_2$ distance of kernel density estimates.

Whereas the bandwidth heuristic used here is fairly crude, much more sophisticated kernel selection procedures exist, e.g.\ by explicitly optimising the test power \citep{Sutherland2017}. A further analysis tool in a similar vein would be to apply a \emph{relative} MMD similarity test \citep{Bounliphone2016}, to rank trained models based on sample fidelity. It would also be possible to adopt a model criticism methodology based on the MMD witness function \citep{Lloyd2015}, to identify over- and under-represented regions in morphometric space (and corresponding generated image exemplars could also be inspected).

\bibliography{references}

\end{document}